\documentclass[a4paper,conference]{IEEEtran}
\usepackage{graphicx}
\usepackage{amsmath}
\usepackage{amssymb}
\usepackage{booktabs}
\usepackage{array}
\usepackage{multirow}
\usepackage{color}
\usepackage{colortbl}
\usepackage{framed}
\usepackage{bm}
\usepackage{bbm}
\usepackage{xspace}
\usepackage{enumitem}
\usepackage[dvipsnames,table]{xcolor}
\usepackage{pifont}
\usepackage{scalerel}
\usepackage{xparse}
\usepackage{xcolor}
\usepackage{lipsum}
\usepackage{url}
\usepackage{tabu}
\usepackage{flushend}

\usepackage[hidelinks]{hyperref}

\def \ie {\emph{i.e.}}
\def \eg {\emph{e.g.}}

\newcommand{\tit}[1]{\smallbreak\noindent\textbf{#1.}}

\ifCLASSINFOpdf
\else
\fi

\begin{document}
\title{The LAM Dataset: A Novel Benchmark for Line-Level Handwritten Text Recognition}

\author{\IEEEauthorblockN{Silvia Cascianelli$^1$, Vittorio Pippi$^1$, Martin Maarand$^2$, Marcella Cornia$^1$, Lorenzo Baraldi$^1$,\\Christopher Kermorvant$^2$, Rita Cucchiara$^1$}
\IEEEauthorblockA{$^1$University of Modena and Reggio Emilia, Modena, Italy \quad \quad $^2$TEKLIA, Paris, France \\
Email: $^1$\{name.surname\}@unimore.it, $^2$\{surname\}@teklia.com}
}

\maketitle

\begin{abstract}
Handwritten Text Recognition (HTR) is an open problem at the intersection of Computer Vision and Natural Language Processing. The main challenges, when dealing with historical manuscripts, are due to the preservation of the paper support, the variability of the handwriting -- even of the same author over a wide time-span -- and the scarcity of data from ancient, poorly represented languages. With the aim of fostering the research on this topic, in this paper we present the Ludovico Antonio Muratori (LAM) dataset, a large line-level HTR dataset of Italian ancient manuscripts edited by a single author over 60 years. The dataset comes in two configurations: a basic splitting and a date-based splitting which takes into account the age of the author. The first setting is intended to study HTR on ancient documents in Italian, while the second focuses on the ability of HTR systems to recognize text written by the same writer in time periods for which training data are not available.
For both configurations, we analyze quantitative and qualitative characteristics, also with respect to other line-level HTR benchmarks, and present the recognition performance of state-of-the-art HTR architectures.
The dataset is available for download at \url{https://aimagelab.ing.unimore.it/go/lam}.

\end{abstract}

\IEEEpeerreviewmaketitle

\section{Introduction}
\label{sec:introduction}
Handwritten Text Recognition (HTR) has been studied for decades~\cite{marti2000handwritten,krevat2006improving,graves2009offline}, thanks to its importance in terms of practical applications (ranging from public administration to industrial processes automation and digital humanities) and for its multimodal and sequential nature, that is common to other pattern recognition tasks. 
Despite the encouraging results achieved by the recent literature, and especially by deep learning-based models~\cite{shi2016end,puigcerver2017multidimensional,yousef2020origaminet,kang2020pay}, HTR is still far from being considered a solved task. 

When performing HTR on historical manuscripts~\cite{rath2007word,bolelli2018xdocs,bouillon2019grayification,santoro2020using,cascianelli2021learning,aradillas2020boosting}, there are additional challenges which need to be taken into account, which are both visual and linguistic issues. From the visual point of view, the digitized images of an historical manuscript exhibit several artifacts: both the paper support and the ink can be deteriorated, and there can be stains, scratches, bleed-through or faded ink. Further, the language used is typically peculiar to the historic period and geographical area in which the manuscript was edited. This often prevents exploiting a pre-trained language model  in modern English language and represents a challenge from the textual and linguistic point of view. Designing effective strategies for the challenges mentioned above requires rich data collections, manually annotated.

\begin{figure}[t]
\centering
\includegraphics[width=\linewidth]{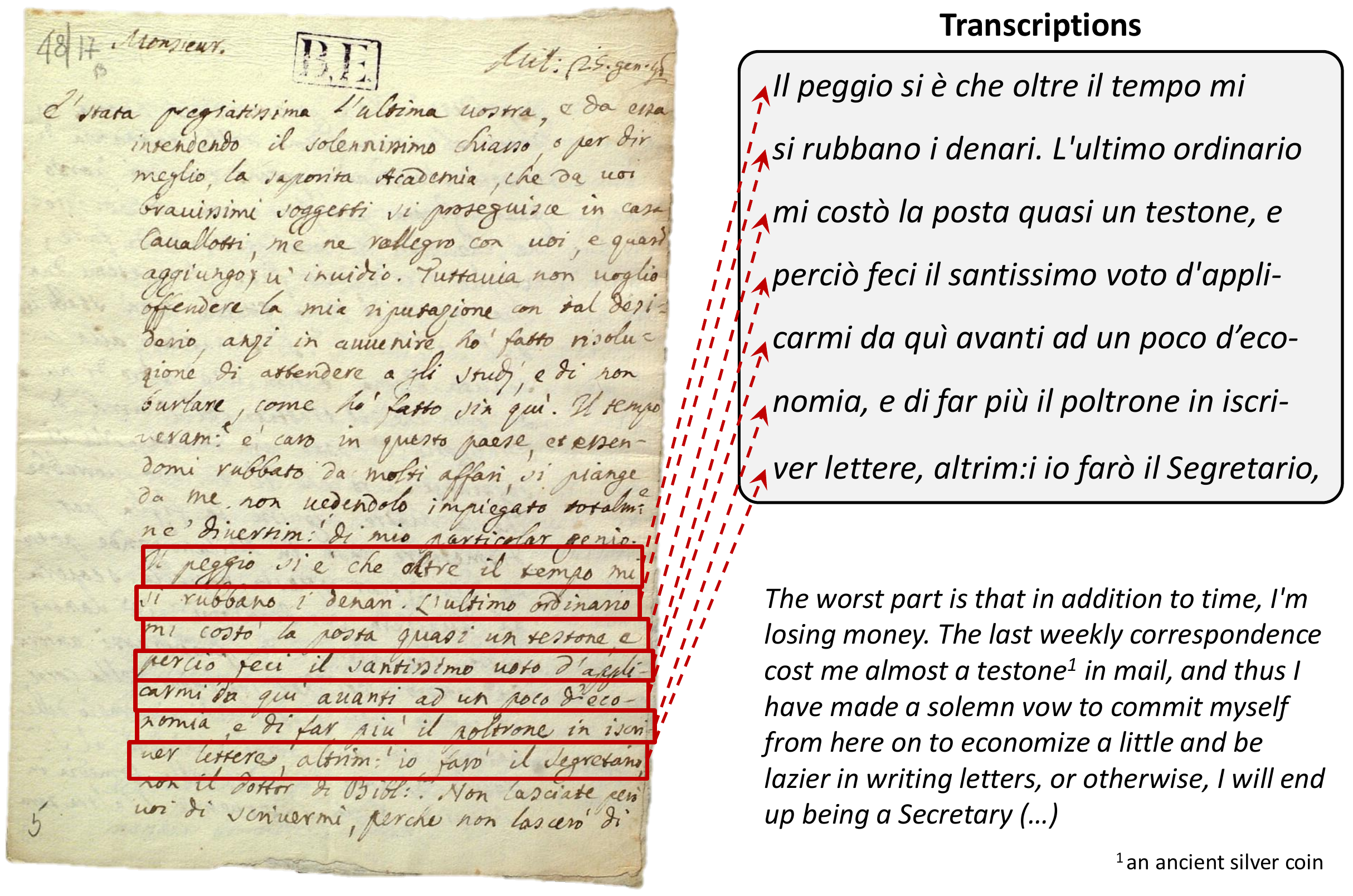}
\caption{The LAM dataset features lines from letters by the Italian historian L. A. Muratori. To the best of our knowledge, it is the largest dataset for line-level HTR.}
\label{fig:first_page}
\vspace{-0.3cm}
\end{figure}

In this paper we contribute to the research on handwriting recognition by presenting a novel and large dataset for HTR on historical manuscripts. 
The dataset is obtained from letters handwritten by the Italian historian Ludovico Antonio Muratori (1672-1750), which are preserved at the Estense Library of Modena (Italy). We selected letters written personally by Muratori (discarding those written by his collaborators, thanks to a precise evaluation by experts), and entirely in ancient Italian\footnote{Muratori, indeed, wrote parts of letters in Spanish, French, and Latin, depending on the correspondent.}. While all letters have been written by a single author, they cover a large time-span of around 60 years -- during which the author's handwriting and paper support varied. This makes the dataset suitable for dealing with the change of handwriting style over time, in addition to being a pure HTR dataset. 
The annotation of the dataset has been performed manually, and it has been double-checked by two experts, who provided diplomatic transcriptions at line-level. 
This level of annotation granularity has been chosen as it is a good trade-off between word-level and paragraph-level in terms of required time, cost, and amount of supervision and because it is common in HTR research. 
Overall, the proposed dataset contains 25,823 lines, which, to the best of our knowledge, makes it \textit{the largest line-level HTR dataset to date}. 

In the reminder of the paper, we present the LAM dataset, named after the initials of Ludovico Antonio Muratori, and provide an overview of its main features by comparing it with existing proposals. Moreover, we perform a quantitative analysis of the performance achieved by different HTR approaches, including both state-of-the-art models and tools, on the dataset with the aim of providing the community with baselines and insights for developing novel architectures for HTR on historical data.

\section{Related Work}
\label{sec:related}
Depending on the choice of the textual unit, HTR can be performed at different granularity levels, ranging from character-level~\cite{jaderberg2015spatial,cilia2019ranking} (which is particularly suited for idiomatic languages) to page-level~\cite{moysset2017full, bluche2017scan, bluche2016joint,wigington2018start,clanuwat2019kuronet}. Line-level HTR is a popular trade-off and one of the most studied variants, especially for non-idiomatic languages. In fact, line-level HTR not only is a standalone variant of the task, but is also often integrated in paragraph-level or page-level HTR systems~\cite{yousef2020origaminet,moysset2017full,bluche2017scan}. For this reason, the LAM dataset has been designed to fit a line-level protocol.

The first approaches to HTR entailed the use of Hidden Markov Models (HMMs), in combination with Gaussian Mixture Models or fully-connected networks for representing the visual input, and \emph{n}-gram based language models for predicting the textual output~\cite{toselli2004integrated,toselli2015handwritten}. In following works, the visual input representation has been performed by using Multi-Dimensional Long Short-Term Memory networks (MD-LSTMs), and the Connectionist Temporal Classifier (CTC) decoding strategy has been introduced to produce the transcription~\cite{bluche2016joint,moysset20192d,de2019no}, as proposed in~\cite{graves2009offline}. Alternatively to MD-LSTMs, CNNs can be used, in combination with one-dimensional LSTMs, to encode the text image~\cite{shi2016end,puigcerver2017multidimensional}. This strategy later became a popular choice~\cite{pham2014dropout, voigtlaender2016handwriting, bluche2017gated, chowdhury2018efficient,cojocaru2020watch,cascianelli2022boosting} since it allows reaching comparable or superior results to MD-LSTM-based approaches, while being faster to train. 
Other approaches have also been investigated to avoid the usage of Recurrent Neural Networks (RNNs). For example, in~\cite{arora2019using}, it is proposed an hybrid approach combining convolutional layers and time-delay neural layers~\cite{waibel1989phoneme} for input representation, with an HMM for output prediction.
Further, in~\cite{yousef2020origaminet,coquenet2020recurrence}, Fully Convolutional Networks (FCNs) are proposed for HTR. To simulate the dependency modeling provided by LSTM cells, FCN are combined with GateBlocks layers~\cite{yousef2018accurate}, which implement a selecting mechanism similar to that of LSTM cells. Each gate is made of Depth-wise Separable Convolutions~\cite{chollet2017xception} to reduce the number of parameters and speed up the training process. 
A recent research line has proposed to apply the sequence-to-sequence paradigm to HTR~\cite{sueiras2018offline, michael2019evaluating}, where the text image is encoded via convolutional and recurrent layers, and the transcription is generated by a RNN-based decoder. As training objective, the CTC loss commonly used in HTR can be combined with the cross-entropy loss.
As a special case of the sequence-to-sequence paradigm, some works apply Transformers~\cite{vaswani2017attention} as encoders or decoders~\cite{wick2021transformer, diaz2021rethinking}, motivated by the success of such architecture in machine translation and language understandings and other vision-and-language tasks~\cite{devlin2018bert,radford2021learning,ramesh2021zero,stefanini2022show}. In HTR, this strategy is effective when sufficient training data is available. For this reason, some works employ synthetic data during a pre-training stage~\cite{kang2020pay,  wick2021rescoring, li2021trocr}. 
Finally, to increase performance, many approaches integrate an explicit language model. This strategy is as effective as the language in the dataset is regular and well-represented, which is not often the case for historical datasets.

\begin{table}[t]
\centering
\small
\setlength{\tabcolsep}{.3em}
\caption{Characteristics of line-level benchmark datasets.}
\label{tab:datasets}
\resizebox{\columnwidth}{!}{%
\begin{tabular}{lc ccccc}
\toprule 
 & & \textbf{Lines} & \textbf{Lexicon} & \textbf{Period} & \textbf{Language} & \textbf{Authors}\\
\midrule
IAM~\cite{marti2002iam} & &  10,373 & 9,749 & Modern & English & Many\\
RIMES~\cite{augustin2006rimes} & &  12,111 & 8,760 & Modern & French & Many\\
\midrule
Washington\cite{fischer2012lexicon} & & 656 & 1,471 & 1755 & English & Two\\
Saint Gall~\cite{fischer2011transcription} & & 1,410 & 5,436 & ca 890-900 & Latin & One\\
Esp. Index~\cite{romero2013esposalles} & & 1,563 & 1,725 & 1491-1495 & Catalan & One\\
Leopardi~\cite{cascianelli2021learning} & &  2,459 & 5,067 & 1818-1832 & Italian & One\\
Parzival~\cite{fischer2012lexicon} & &  4,477 & 4,934 & ca 1265-1300 & German & Two\\
Esp. Licenses~\cite{romero2013esposalles} & & 5,447 & 3,465 & 1616-1619 & Catalan & One\\
ICFHR14~\cite{sanchez2014icfhr2014} & &  11,473 & 9,716 & ca 1760-1832 & English & Many\\
ICFHR16~\cite{sanchez2016icfhr2016} & &  10,550 & 8,120 & 1470-1805 & German & Many\\
ICFHR18~\cite{strauss2018icfhr2018} & &  14,803 & 23,198 & Mixed\footnotemark[1] & German, Italian & Many\\
Rodrigo~\cite{serrano2010rodrigo} & &  20,357 & 17,300 & 1545 & Spanish & One\\
Germana~\cite{perez2009germana} & &  20,529 & 27,100 & 1891 & Spanish\footnotemark[2] & One\\
\midrule
\textbf{LAM (Ours)} & & 25,823 & 23,428 & 1691-1750 & Italian & One\\
\bottomrule
\addlinespace[0.1cm]
\multicolumn{7}{l}{\footnotesize 1 - Both Medieval and Modern.} \\
\multicolumn{7}{l}{\footnotesize 2 - A small number of lines are in different languages.}
\end{tabular}
}\\
\vspace{-.3cm}
\end{table}

\tit{Existing Benchmark Datasets}
Designing and developing effective HTR solutions requires the availability of large data collections, which should capture both the visual variability of the task and represent different languages. In the following, we focus on line-level dataset of western-characters, since these are more closely related to our proposed dataset. The main characteristics of those datasets are also reported in Table~\ref{tab:datasets}.

Commonly-used benchmarks for line-level HTR include the IAM~\cite{marti2002iam} and the RIMES~\cite{augustin2006rimes} datasets, both containing lines in modern languages (English and French, respectively) and written by multiple authors on regular paper support. The language used is somewhat constrained, since the writers have been carefully instructed on what to write: copying English sentences from the Lancaster-Oslo/Berge corpus~\cite{johansson1978manual} in IAM, and following a template and a script for writing customer service-themed letters in RIMES. 

As for HTR on historical manuscripts, many datasets have been released to explore different perspectives of the task. Among those, the most used are the ICFHR14~\cite{sanchez2014icfhr2014}, ICFHR16~\cite{sanchez2016icfhr2016}, and the ICFHR18~\cite{strauss2018icfhr2018} datasets, all prepared for HTR challenges at the International Conference on Frontiers of Handwriting Recognition (ICFHR). The aim of ICFHR14 is to explore HTR on historical data rather than modern ones. Therefore, the dataset contains lines form the Bentham Papers collection~\cite{causer2012building}, handwritten in English by few authors, mainly the philosopher Jeremy Bentham and his collaborators. The ICFHR16 was initially intended to study HTR on a language that is structurally different from English, and thus it contains lines from the Ratsprotokolle collection, handwritten in German by multiple writers in over three centuries. Finally, the ICFHR18 dataset was designed to investigate the minimum amount of training data required to correctly transcribe an entire historical document. For this reason, the dataset contains documents from many different collections and time periods, it is written in different languages (German and Italian), and its test set is divided in document-specific sets. Moreover, it is worth mentioning the Rodrigo~\cite{serrano2010rodrigo} and Germana~\cite{perez2009germana} datasets. These are large line-level datasets obtained from two different Spanish books %(although the Germana datasets also contains some lines in different languages) 
and written by a single author in a short time-span.

There are also other datasets containing historical manuscripts, which are of much smaller size and thus can be used to explore HTR in the case of limited training data and specific domain. Examples of such datasets are the George Washington dataset~\cite{fischer2012lexicon}, containing English letters by George Washington (and few parts by a collaborator), and the Parzival dataset~\cite{fischer2011transcription}, containing a Medieval German poem handwritten by two writers. Usually, datasets of this kind feature documents handwritten by a single author in a relatively limited time-span, during which the handwriting does not change significantly. Some examples are the Saint Gall dataset~\cite{fischer2009automatic}, with lines from a Medieval Latin manuscript, the Esposalles Index and Esposalles Licenses datasets~\cite{romero2013esposalles}, with lines from Catalan marriages registers, and the Leopardi dataset~\cite{cascianelli2021learning}, with Italian letters by the writer Giacomo Leopardi.

\section{The LAM Dataset}
\label{sec:dataset}
In this section, we analyze the main characteristics of the proposed dataset. It comes with different splittings to allow performing classical HTR, on a splitting we refer to as \emph{basic split}, and time-dependent HTR, in a setting we refer to as \emph{date-based setting}. The main characteristics of these splittings are reported in Table~\ref{tab:splittings}.

\subsection{Data Collection and Preparation}
The documents used for the LAM dataset come from the digitized L. A. Muratori collection preserved at the Estense Digital Library. The collection contains drafts, papers, notes, and letters handwritten by the Italian historian and his collaborators. Some of these documents, or parts of those, are written in languages different from Italian, which include Latin, French, and Spanish.

For collecting the dataset, we prepared an ad-hoc on-line annotation tool. We preferred not to use available commercial tools to obtain simplicity of use by non-experts, to keep the data in house before the release of this dataset, and to favor crowd-sourcing since the tool does not require any license or subscription to be used. 
Further details on the developed platform can be found in the supplementary material.

Two experts were involved in the data preparation. 
First, they selected from the considered collection of digitized documents, only autograph letters by Muratori in Italian, for a total of 1,171 pages from 72 files edited in a time-span of 60 years. 
Then, they annotated the letters at line-level, by providing the bounding box of each line and its diplomatic transcription. In the transcription process, stroke-out text, words that are illegible due to stains and scratches, and special symbols not representable in Unicode have been replaced with ``\texttt{\#}''. 

Note that each considered file contains letters to a different correspondent, which was either a family member, a friend, a professional or an acquaintance of different social extraction and cultural level, to whom Muratori wrote about many different topics. This results in a rich and varied language. 
The annotation process and subsequent double-checking took approximately one year and, to the best of our knowledge, lead to the largest dataset for line-level HTR to date.

\begin{table}[t]
\centering
\small
\setlength{\tabcolsep}{.3em}
\caption{LAM dataset splits statistics. The charset size is calculated on the Training and Validation splits.}
\label{tab:splittings}
\resizebox{0.99\columnwidth}{!}{%
\begin{tabu}{lc ccccc}
\toprule 
 & & \textbf{Total} & \textbf{Training} & \textbf{Validation} & \textbf{Test} & \textbf{Charset} \\
\midrule
\textbf{Basic split}  & & 25,823 & 19,830 & 2,470 & 3,523 & 89\\
\textbf{Date-based setting (leave-decade-out)} \\
\hspace{1.125cm}\textit{1690-1700} & & 25,183 & 17,205 & 1,911 & 6,067 & 87\\
\hspace{1.125cm}\textit{1700-1710} & & 22,392 & 17,205 & 1,911 & 3,276 & 84\\
\hspace{1.125cm}\textit{1710-1720} & & 21,066 & 17,205 & 1,911 & 1,950 & 83\\
\hspace{1.125cm}\textit{1720-1730} & & 25,158 & 17,205 & 1,911 & 6,042 & 86\\
\hspace{1.125cm}\textit{1730-1740} & & 22,974 & 17,205 & 1,911 & 3,858 & 85\\
\hspace{1.125cm}\textit{1740-1750} & & 23,106 & 17,205 & 1,911 & 3,990 & 84\\
\textbf{Date-based setting (decade-vs-decade)} \\
\hspace{1.125cm}\textit{1690-1700} & & 25,183 & 5,460 & 607 & 19,116 & 80\\
\hspace{1.125cm}\textit{1700-1710} & & 25,183 & 2,948 & 328 & 21,907 & 80\\
\hspace{1.125cm}\textit{1710-1720} & & 25,183 & 1,755 & 195 & 23,233 & 77\\
\hspace{1.125cm}\textit{1720-1730} & & 25,183 & 5,437 & 605 & 19,141 & 81\\
\hspace{1.125cm}\textit{1730-1740} & & 25,183 & 3,472 & 386 & 21,325 & 83\\
\hspace{1.125cm}\textit{1740-1750} & & 25,183 & 3,591 & 399 & 21,193 & 81\\
\bottomrule
\end{tabu}
}
\vspace{-.3cm}
\end{table}

\subsection{General Characteristics}
The LAM dataset contains a total of 25,823 lines, with a lexicon of over 23,000 unique words (see Table~\ref{tab:datasets}). Other datasets of comparable size feature a rich lexicon as well, especially those containing text in different languages (\eg~the ICFHR18 and Germana datasets). In the case of the LAM dataset, the richness of the lexicon is due to the fact that, in his letters, Muratori wrote about different topics, mentioned many different proper nouns (of people and places), and used various forms of abbreviations for names, titles, and salutations, which was common in his time to save paper when writing. 

As for the visual characteristics, the LAM dataset features all the typical time-related challenges of historical manuscripts. In particular, the paper support used varies considerably from page to page (both in terms of color and texture), and there are pages with humidity stains, creases, scratches, and holes. Also the ink used makes the dataset challenging since there are pages with faded or bled trough ink, stains, discolorations, and corrosions. Some pages from the LAM dataset exemplifying the aforementioned challenges are shown in the supplementary material.

\begin{figure}[t]
\centering
\includegraphics[width=\columnwidth]{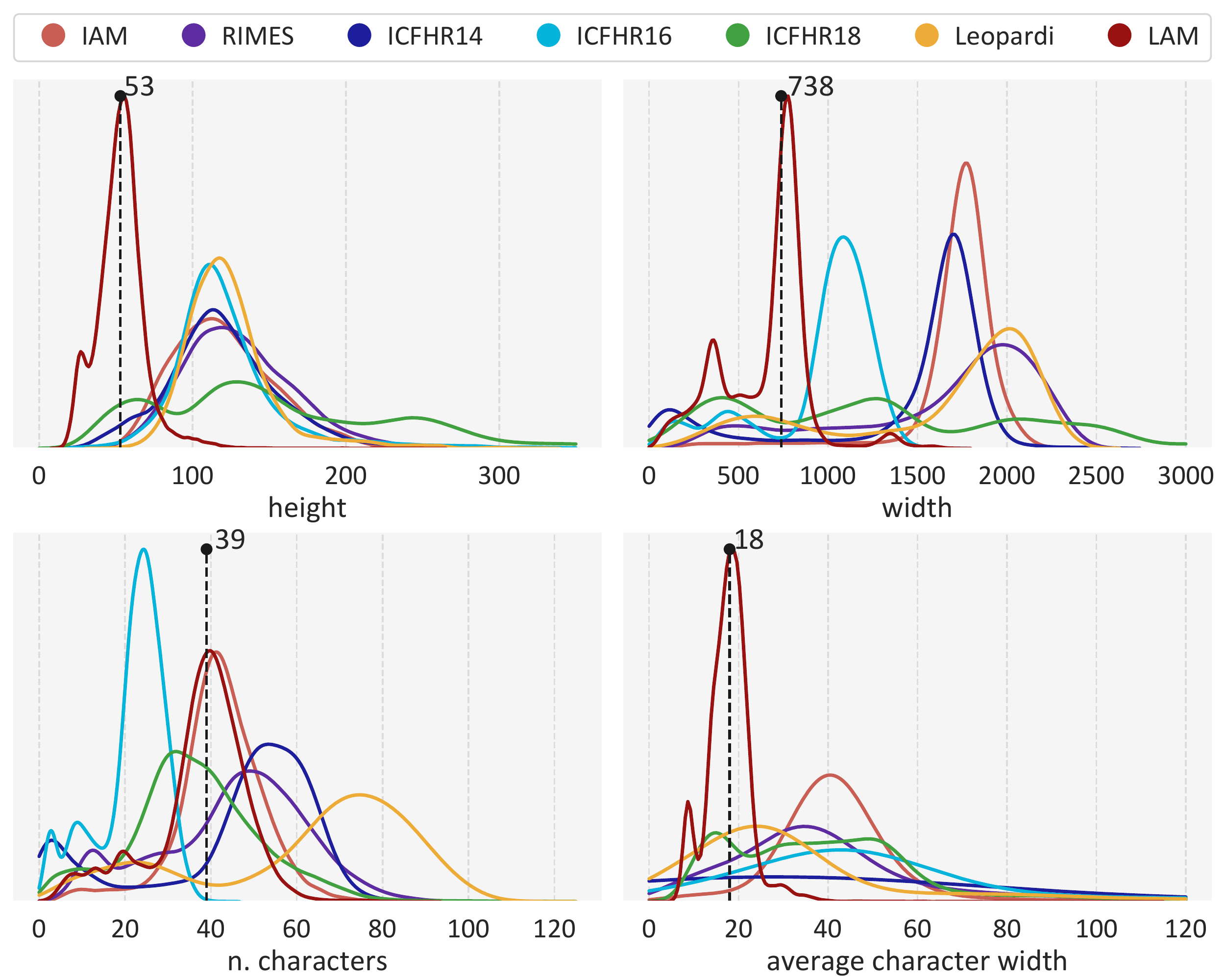} 
\caption{Lines height and width distribution in the LAM dataset compared to other benchmark datasets (top). Number of characters per line distribution in the LAM dataset compared to other popular benchmark datasets (bottom left). Average characters pixel width distribution in the LAM dataset compared to other benchmark datasets (bottom right). Best seen in color.}
\label{fig:datasets_comparison}
\vspace{-.5cm}
\end{figure}

Further characteristics of the LAM dataset are analyzed in Fig.~\ref{fig:datasets_comparison} in comparison with other commonly used benchmark datasets, both modern and historical, and with a smaller historical dataset in Italian (the Leopardi dataset). Compared to other datasets, the lines in LAM have smaller and more regular height, while the line images width has a more clear tendency to bimodality. This is due to the fact that, depending on the content and the addressee, some pages are written in double-column. Moreover, to save paper, the author commonly exploited the entire text column width, which also explains the clearer regularity in the width of the characters compared to other datasets. Also in terms of the average number of characters per line LAM shows regularity, having the majority of lines with 39 characters (similar to IAM).

To use the LAM dataset for classic HTR, we provide a \textbf{basic split} consisting of 19,830 lines for training, 2,470 for validation, and 3,523 for test. The lines have been collected from different portions of the pages in each of the 72 considered files, of 80\%, 10\%, and 10\%, respectively. This splitting is intended for exploring HTR on images featuring the typical challenges of historical manuscripts and a rich and underrepresented language such as ancient Italian.

\subsection{Date-based Setting}\label{ssec:date_based_split}
As mentioned above, the documents from which we collected the LAM dataset cover roughly 60 years. 
For most of the letters in the files the date in which they have been written is clearly indicated. Therefore, according to this information, we were able to separate them into six groups reflecting the decade. 
The idea behind this date-based setting is to explore the effect of the availability of handwritten samples from an author in different time periods over the recognition of his/her text in an unseen time period. In a wide time-span as that considered in the LAM dataset, the handwriting of the author is likely to change. 
In this respect, a t-SNE analysis of the lines in the date-based setting and examples of pages from the six splits can be found in the supplementary material.

After discarding pages with no date indication (27 out of 1,171 pages), we built two setups that can be used to perform HTR of the same author, conditioned on time. %
In the first setup, referred to as \textbf{leave-decade-out}, the test set contains all the lines from the pages of a certain decade, while the training and validation sets contain a proportional amount of lines from the pages of the other decades (90\% and 10\%). Note that, for fair comparison and data balancing, we include the same amount of lines in the training set and the validation set of each split. In the second setup, referred to as \textbf{decade-vs-decade}, all the lines from a decade of choice are used for training and validation, and all the lines from each other decade separately are used for test.
The size of the subsets and the charset in each date-based split is reported in Table~\ref{tab:splittings}.

\section{Experimental Evaluation}
\label{sec:experiments}
In this section, we report an experimental analysis of the performance of popular state-of-the-art models and toolkits, both on the basic split of the LAM dataset and on the date-based setting. The performance are reported in terms of Character Error Rate (CER) and Word Error Rate (WER). As customary in HTR, to calculate the CER and the WER on the entire test set, we first compute the Edit Distance (at character level for the CER and word level for the WER) between each predicted sentence and the corresponding ground truth. This is the number of substitutions, deletions, and insertions that have to be applied to the predicted sentence to obtain the ground truth. Then, we sum up the distances of all samples, divide by the sum of the ground truth lengths and multiply by 100.

\subsection{Considered HTR Approaches}
We consider models following different kinds of architectures for HTR in order to give insights on the possibly more promising strategy to be applied on the LAM dataset and guide future research. 
When available, we used the official implementation and weights provided by the authors, while in the other cases, we used our best implementation. All the models have been trained by following the training protocol described in the original paper.  
Note that data augmentation is not performed in any of the considered approaches to better highlight the effect of the size of the LAM dataset and the data variability it captures. For methods requiring an explicit charset, this has been obtained from the training and validation subsets of the basic split, and of each of the splittings in the date-based setting (see Table~\ref{tab:splittings}). 

\subsubsection{Convolutional-Recurrent Paradigm} 
Combining CNNs and RNNs for HTR has been the standard choice for years. In this analysis we consider models featuring 1D-LSTMs, since these have been proven to be comparable or superior to MDLSTMs~\cite{graves2009offline} while being much faster to train~\cite{puigcerver2017multidimensional}.
In particular, we test our implementation of the \textbf{1D-LSTM} proposed in~\cite{puigcerver2017multidimensional}, which consists of a stack of five convolutional blocks and five Bidirectional Long Short-Term Memory network (BLSTM) layers. We also consider the approach proposed in~\cite{shi2016end} (referred to as \textbf{CRNN} in the following), which has a deeper convolutional component but fewer recurrent layers compared to 1D-LSTM. Specifically, in this variant there are seven convolutional blocks, two of whom contain rectangular max-pooling layers to better maintain the aspect ratio of the text lines, and two BLSTM layers. For both the 1D-LSTM and CRNN approaches, we additionally consider variants containing Deformable Convolutions (DefConvs)~\cite{dai2017deformable}, as proposed in~\cite{cascianelli2021learning,cojocaru2020watch}. Finally, we include in the analysis the default model available from the popular HTR toolkit \textbf{PyLaia}~\cite{puigcerver2018}, which has four convolutional layers and three BLSTM layers. 

\subsubsection{Sequence-to-Sequence Paradigm with Transformers}
As a representative of the sequence-to-sequence paradigm, we explore Transformer-based approaches, which are more data-demanding than classical RNN-based solutions. Therefore, by considering these kind of architectures, we can investigate whether the size of the LAM dataset allows effectively training large HTR models. In this respect, we consider the strategy proposed in~\cite{kang2020pay} (referred to as \textbf{Transformer} in the following). This architecture exploits a ResNet-101 trained from scratch and a Transformer encoder and decoder~\cite{vaswani2017attention} with reduced parameters. The ResNet produces a feature map that is then flattened and used as input embeddings for the Transformer architecture. 
Moreover, we consider the Base version of the  \textbf{TrOCR} model proposed in~\cite{li2021trocr}, which employs Transformer-like architectures both for image representation~\cite{dosovitskiy2021image,touvron2021training} and text generation~\cite{vaswani2017attention,devlin2018bert}, and exploits large-scale pre-training, both on typewritten and handwritten lines, before being fine-tuned on the dataset of interest.

\subsubsection{Fully-Convolutional Paradigm}
Recent approaches to HTR come with FCNs, establishing state-of-the-art results. 
In this work, we consider the Gated Fully Convolutional Network \textbf{(GFCN)}~\cite{coquenet2020recurrence}, which preprocesses the input image with four convolutional layers and then passes the output through five GateBlocks layers~\cite{yousef2018accurate}. 
We also consider three different variants of the deeper model \textbf{OrigamiNet}~\cite{yousef2020origaminet}, containing 12, 18, and 24 GateBlocks layers, respectively. Moreover, we include in the analysis the implementation of the approach proposed in~\cite{arora2019using} available in the \textbf{Kaldi} toolkit. This model is composed of six convolutional layers and three time-delay neural layers, followed by an HMM for text recognition. The architecture training is divided into two phases. First, a ``flat start'' model is trained on images and the corresponding transcriptions. Then, the trained ``flat start'' model is used to create alignments for training a second model, which is the only one used at inference time. Finally, a 3-gram byte pair encoding language model is applied to improve the decoding.

\begin{table}[t]
\centering
\small
\setlength{\tabcolsep}{.25em}
\caption{Results on the test sets of the LAM basic split, and on the IAM and the ICFHR14 datasets. The $^*$ marker indicates our re-implementations.}
\label{tab:results}
\resizebox{\linewidth}{!}{%
\begin{tabu}{lc c cc c cc c cc}
\toprule 
& & & \multicolumn{2}{c}{\textbf{IAM}} & & \multicolumn{2}{c}{\textbf{ICFHR14}} & & \multicolumn{2}{c}{\textbf{LAM}} \\
\cmidrule{4-5} \cmidrule{7-8} \cmidrule{10-11}
\textbf{Method} & \textbf{\#Params (M)} & & \textbf{CER} & \textbf{WER} & & \textbf{CER} & \textbf{WER} & & \textbf{CER} & \textbf{WER} \\
\midrule 
\textit{HTR Toolkits} \\
\hspace{0.4cm}\textbf{PyLaia~\cite{puigcerver2018}}  & 4.8 & & 9.8 & 32.3 & & 5.1 & 17.5 & & 4.7 & 16.5  \\
\hspace{0.4cm}\textbf{Kaldi~\cite{arora2019using}}   & 15.0  & &  7.2 & 25.0 & & 3.7 & 14.2 & & 4.7 & 13.4  \\
\midrule
\textit{HTR Models} \\

\hspace{0.4cm}\textbf{1D-LSTM~\cite{puigcerver2017multidimensional}}        & -   & & 8.3   & 24.9 & & -   & -  & & -   & - \\
\hspace{0.4cm}\textbf{1D-LSTM~\cite{puigcerver2017multidimensional}}$^*$        & 9.6   & & 7.7   & 26.3 & & 4.8   & 15.3  & & 3.7   & 12.3 \\
\hspace{0.4cm}\textbf{1D-LSTM (w/ DefConv)~\cite{cojocaru2020watch}}            & 9.6   & & 7.5   & 26.9 & & 3.6   & 14.3  & & 3.5   & 11.6 \\
\hspace{0.4cm}\textbf{CRNN~\cite{shi2016end}}$^*$                               & 18.2  & & 7.8   & 27.8 & & 3.9   & 15.3  & & 3.8   & 12.9 \\
\hspace{0.4cm}\textbf{CRNN (w/ DefConv)~\cite{cojocaru2020watch}}               & 18.5  & & 6.8   & 24.7 & & 3.6   & 13.9  & & 3.3   & 11.3 \\
\hspace{0.4cm}\textbf{Transformer~\cite{kang2020pay}}$^*$                       & 54.7  & & -     & -    & & -     & -     & & 10.2  & 22.0 \\

\hspace{0.4cm}\textbf{TrOCR~\cite{li2021trocr}}            & 385.0 & & 3.4   & - & & -   & -  & & -   & - \\
\hspace{0.4cm}\textbf{TrOCR~\cite{li2021trocr}}$^*$                       & 385.0 & & 7.3   & 37.5 & & 3.5   & 11.5  & & 3.6   & 11.6 \\
\hspace{0.4cm}\textbf{GFCN~\cite{coquenet2020recurrence}}  & 1.4   & & 8.0   & 28.6 & & -   & -  & & -   & - \\
\hspace{0.4cm}\textbf{GFCN~\cite{coquenet2020recurrence}}$^*$                       & 1.4   & & 8.0   & 28.6 & & -   & -  & & 5.2 & 18.5 \\
\hspace{0.4cm}\textbf{OrigamiNet$_{\text{12}}$~\cite{yousef2020origaminet}} & 39.0 & & 5.3   & - & & -   & -  & & -   & - \\
\hspace{0.4cm}\textbf{OrigamiNet$_{\text{12}}$~\cite{yousef2020origaminet}}$^*$     & 39.0  & & 6.0   & 22.3 & & 3.6   & 14.7  & & 3.1   & 11.2 \\
\hspace{0.4cm}\textbf{OrigamiNet$_{\text{18}}$~\cite{yousef2020origaminet}} & 77.1 & & 4.8   & - & & -   & -  & & -   & - \\
\hspace{0.4cm}\textbf{OrigamiNet$_{\text{18}}$~\cite{yousef2020origaminet}}$^*$     & 77.1  & & 6.6   & 24.2 & & 4.0   & 15.4  & & 3.1   & 11.0 \\
\hspace{0.4cm}\textbf{OrigamiNet$_{\text{24}}$~\cite{yousef2020origaminet}} & 115.3 & & 4.8   & - & & -   & -  & & -   & - \\
\hspace{0.4cm}\textbf{OrigamiNet$_{\text{24}}$~\cite{yousef2020origaminet}}$^*$     & 115.3 & & 6.5   & 23.9 & & 5.9   & 21.3  & & 3.0   & 11.0 \\
\bottomrule
\end{tabu}
}
\vspace{-.2cm}
\end{table}

\begin{table*}[t]
\centering
\small
\setlength{\tabcolsep}{.4em}
\caption{Results on the leave-decade-out setup of the date-based setting. The $^*$ marker indicates our re-implementations.}
\label{tab:results_date}
\resizebox{0.85\textwidth}{!}{%
\begin{tabular}{lc cc c cc c cc c cc c cc c cc c cc}
\toprule 
& & \multicolumn{2}{c}{\textbf{1690-1700}} & & \multicolumn{2}{c}{\textbf{1700-1710}} & & \multicolumn{2}{c}{\textbf{1710-1720}}  & & \multicolumn{2}{c}{\textbf{1720-1730}} & & \multicolumn{2}{c}{\textbf{1730-1740}} & & \multicolumn{2}{c}{\textbf{1740-1750}} & & \multicolumn{2}{c}{\textbf{Average}}\\
\cmidrule{3-4} \cmidrule{6-7} \cmidrule{9-10} \cmidrule{12-13} \cmidrule{15-16} \cmidrule{18-19} \cmidrule{21-22}
\textbf{Method} & & \textbf{CER} & \textbf{WER} & & \textbf{CER} & \textbf{WER} & & \textbf{CER} & \textbf{WER} & & \textbf{CER} & \textbf{WER} & & \textbf{CER} & \textbf{WER} & & \textbf{CER} & \textbf{WER} & & \textbf{CER} & \textbf{WER}\\
\midrule
\textit{HTR Toolkits} \\ 
\hspace{0.4cm}\textbf{PyLaia~\cite{puigcerver2018}}  & & 6.0 & 23.3 & & 3.7 & 13.7 & & 3.1 & 11.2 & & 3.0 & 11.5 & & 4.8 & 16.1 & & 3.9 & 14.5 & & 4.0 & 15.1 \\
\hspace{0.4cm}\textbf{Kaldi~\cite{arora2019using}}   & & 4.9 & 19.1 & & 3.0 & 10.4 & & 2.7 & 9.7 & & 2.5 & 9.4 & & 4.5 & 13.1 & & 3.2 & 11.4 & &    3.5 & 12.2 \\
\midrule
\textit{HTR Models} \\
\hspace{0.4cm}\textbf{1D-LSTM~\cite{puigcerver2017multidimensional}}$^*$    & & 5.3  & 20.9 & & 3.7  & 12.6 & & 2.8  & 9.1  & & 2.7  & 9.0  & & 4.1  & 14.3 & & 3.6  & 11.8 & & 3.7  & 13.0\\
\hspace{0.4cm}\textbf{1D-LSTM (w/ DefConv)~\cite{cojocaru2020watch}}        & & 5.0  & 19.8 & & 3.6  & 12.1 & & 2.5  & 8.3  & & 3.1  & 10.1 & & 3.8  & 12.3 & & 3.6  & 12.2 & & 3.6  & 12.5\\
\hspace{0.4cm}\textbf{CRNN~\cite{shi2016end}}$^*$                           & & 5.0  & 20.1 & & 3.5  & 12.1 & & 2.6  & 8.8  & & 3.1  & 10.4 & & 4.3  & 14.0 & & 3.8  & 12.7 & & 3.7  & 13.0\\
\hspace{0.4cm}\textbf{CRNN (w/ DefConv)~\cite{cojocaru2020watch}}           & & 4.7  & 19.0 & & 3.3  & 11.1 & & 2.2  & 7.6  & & 2.4  & 8.3  & & 3.7  & 12.2 & & 3.4  & 11.1 & & 3.3  & 11.6\\
\hspace{0.4cm}\textbf{Transformer~\cite{kang2020pay}}$^*$                   & & 15.2 & 37.2 & & 18.6 & 36.5 & & 14.6 & 28.5 & & 10.0 & 20.1 & & 19.5 & 35.7 & & 11.9 & 25.2 & & 15.0 & 30.5\\  
\hspace{0.4cm}\textbf{GFCN~\cite{coquenet2020recurrence}}                   & & 5.1  & 18.5 & & 7.4  & 23.1 & & 3.0  & 10.8 & & 4.2  & 14.6 & & 5.5  & 17.7 & & 4.2  & 15.4 & & 4.9  & 16.7\\
\hspace{0.4cm}\textbf{OrigamiNet$_{\text{12}}$~\cite{yousef2020origaminet}} & & 4.6  & 18.9 & & 2.8  & 10.3 & & 2.2  & 8.0  & & 2.2  & 8.3  & & 3.4  & 11.4 & & 3.0  & 11.8 & & 3.0  & 11.5\\
\hspace{0.4cm}\textbf{OrigamiNet$_{\text{18}}$~\cite{yousef2020origaminet}} & & 4.5  & 18.7 & & 2.8  & 10.3 & & 2.2  & 8.1  & & 2.3  & 8.9  & & 3.5  & 11.8 & & 2.3  & 8.9  & & 2.9  & 11.1\\
\hspace{0.4cm}\textbf{OrigamiNet$_{\text{24}}$~\cite{yousef2020origaminet}} & & 4.9  & 20.4 & & 2.9  & 10.6 & & 2.3  & 8.2  & & 2.2  & 8.4  & & 3.3  & 10.9 & & 3.1  & 11.6 & & 3.1  
& 11.7\\
\bottomrule
\end{tabular}
}
\vspace{-.2cm}
\end{table*}

\subsection{Evaluation Results}
\tit{LAM Basic Split}
The results obtained by the models included in the analysis on the basic split are reported in Table~\ref{tab:results}. 
In this setting, the models following the convolutional-recurrent paradigm obtain a CER on average below 4\% and a WER below 12\%. The performance improvement given by the use of DefConvs indicates the importance of a strong image encoder in this dataset. Among the approaches following the fully-convolutional paradigm, only OrigamiNet performs on par with the convolutional-recurrent approaches, and even performs best in its variant containing 24 GateBlocks. This can be traced back to the high number of GateBlocks that this model contains, allowing to better model the context compared to GFCN and Kaldi, confirming the importance of the image representation for this dataset. Another observation comes from the relatively poor performance of the Transformer model. This implements an intrinsic language model that is challenged by the heavy use of rare words and abbreviations in this dataset. The TrOCR model, which features both a strong image representation component and an intrinsic language model pre-trained on a large amount of image-text pairs, reaches error rates that are comparable with those of the convolutional-recurrent models after being fine-tuned for 30 epochs on the LAM dataset.

\begin{figure}[t]
\centering
\footnotesize
\setlength{\tabcolsep}{.7em}
\resizebox{\linewidth}{!}{
\begin{tabular}{ll}
\multicolumn{2}{c}{\includegraphics[width=0.8\linewidth]{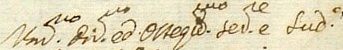}} \\
\textbf{Ground-truth}         & Um.mo Div.mo ed Osseq.mo Ser.re e Sud.o \\
\textbf{CRNN (w/ DefConv)}    & Vo.mo div.mo ed Ossegl.mo Ser.r e Sud.o  \\
\textbf{TrOCR}                & Um.mo Div.mo ed ossequ.mo Ser.re e Sud.o \\
\textbf{OrigamiNet$_{\text{24}}$} & Um.mo Di.o ed Ossegl.mo Ser.r e Sud.e \\
\\
 \multicolumn{2}{c}{\includegraphics[width=0.95\linewidth]{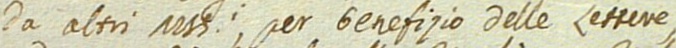}} \\
\textbf{Ground-truth}         & da altri \# per benefizio delle Lettere, \\
\textbf{CRNN (w/ DefConv)}    & da altri MSS.i, per benefizio delle Lettere, \\
\textbf{TrOCR}                & da altri Mess.i, per benefizio delle Lettere \\
\textbf{OrigamiNet$_{\text{24}}$} & da altri MS\#.i, per benefizio delle Lettere \\
\end{tabular}
}
\caption{Qualitative results of the best performing models on example challenging lines from the LAM dataset.}
\label{fig:qualitatives_few}
\vspace{-.5cm}
\end{figure}

As a further comparison between the LAM dataset and other existing benchmarks, we evaluate the performance of the considered models also on IAM and ICFHR14. A performance drop can be noticed for all models, especially bigger ones, some of which did not converge in some cases (\ie~Transformer and GFCN). This indicates that the large amount of training data provided by the LAM dataset can contribute to enable the development of effective models for HTR, similar to what has been done for other vision-and-language tasks.

Finally, we report the qualitative results of the best performing models following the three considered HTR paradigms (CRNN with DefConv, TrOCR, and OrigamiNet$_{\text{24}}$) on challenging lines of the LAM dataset in Fig.~\ref{fig:qualitatives_few}. Additional examples can be found in the supplementary material.

\tit{LAM Date-based Setting}
As for the leave-decade-out setup of the date-based setting, we report the results of the considered approaches in Table~\ref{tab:results_date}. For all the models, the worst performance is obtained on the fifth-decade split (1730-1740). The first and second splits are challenging as well, considering the performance of GFCN and Transformer on the second decade, and of all the other approaches on the first decade. 
The splits having the third and fourth decade as test set are instead easier to recognize. In fact, the errors of the considered approaches on these splits are even lower than what obtained on the basic split. Arguably, this is due to a more homogeneous and clear handwriting of the author in his middle age. 
Additional to a decade-specific analysis, to express the overall performance of HTR models to recognize text over time, we propose to use the average CER and WER on the six splits. According to these scores, the best-performing model in this setting is OrigamiNet in its variant with 18 GateBlocks.

\begin{figure}
    \centering
    \includegraphics[width=\columnwidth]{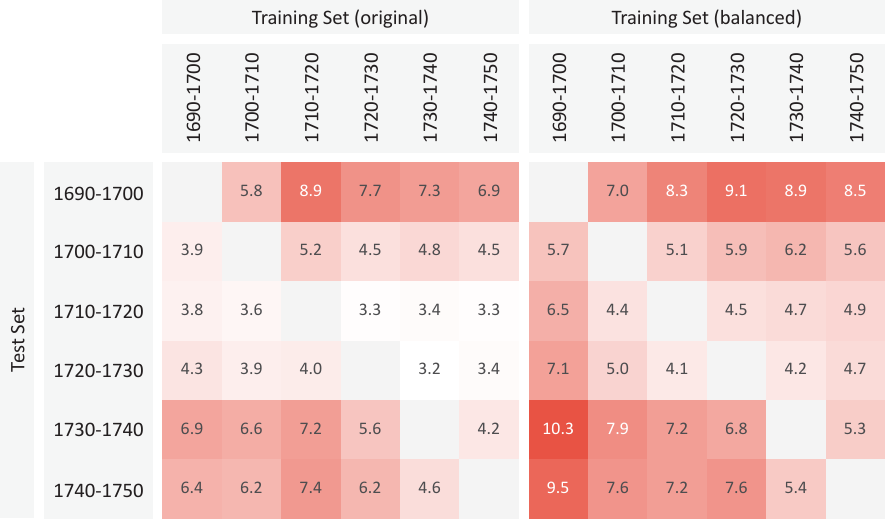}
    \caption{Results of OrigamiNet$_{\text{18}}$ in the decade-vs-decade setup of the date-based setting, both in the scenario in which all the training samples available for each decade are used (left) and the balanced scenario (right).}
    \label{fig:dec_vs_dec}
    \vspace{-.5cm}
\end{figure}

To further explore the challenges posed by the date-based setting, we consider OrigamiNet$_{\text{18}}$ in the decade-vs-decade setup. The CER values obtained in this experiment are reported in Fig.~\ref{fig:dec_vs_dec} (the WER scores are reported in the supplementary material). Overall, the first and the last two decades are the most challenging to recognize, while the text produced in the author's middle age is easier to recognize. The results reported in the table also highlight the difficulty in transcribing documents written at an early age when training on those written at a late age and vice versa, which is a challenge posed by the date-based setting. 
Moreover, to assess whether the difference in performance can be attributed to the difference in the number of samples available for each decade, we repeat the above analysis by using training sets of equal size (artificially balanced by randomly sampling an equal number of lines for each decade) and the same test sets as in Table~\ref{tab:splittings} (further details are given in the supplementary material). The results of this analysis are reported in Fig.~\ref{fig:dec_vs_dec}. Despite the expected numerical variations in the specific CER values, the same observations made in the case of the released setup apply also in this case of balanced setup, thus allowing imputing the challenges emerged to the characteristics of the data rather than to the training set size.

\section{Conclusion}
\label{sec:conclusion}
In this work, we presented the LAM dataset for line-level HTR on historical manuscripts, containing more than 25,000 lines. The dataset features letters written in Italian by a single author over around 60 years, which makes it suitable not only for research on HTR, but also on handwriting recognition over time. To this end, the dataset comes with different splits, reflecting the decade in which the letters have been written. Quantitative and qualitative analyses of the dataset, both of its characteristics and performance achievable with commonly used HTR approaches, highlight the challenges posed by the LAM dataset, which we hope can make it a valuable contribution towards the development of effective solutions to HTR on historical documents. 
As a further development of this work, the LAM dataset could be enriched with word-level annotations, thus increasing the level of supervision and making it suitable also for the keyword spotting task on historical manuscripts.

\small{\section*{Acknowledgment}
This work was supported by the ``AI for Digital Humanities'' project, funded by ``Fondazione di Modena'', by the ``DHMoRe Lab'' project, funded by ``Regione Emilia Romagna'', and by the ``Artificial Intelligence for Cultural Heritage'' project, co-funded by the Italian Ministry of Foreign Affairs and International Cooperation. 
The authors thank Dr. Maria Ludovica Piazzi, Dr. Rosiana Schiuma, and the Estense Digital Library for the contribution and support provided in preparing the dataset.
}

\clearpage
\bibliographystyle{IEEEtran}
\bibliography{bibliography}

\end{document}

\title{The LAM Dataset: A Novel Benchmark for Line-Level Handwritten Text Recognition\\Supplementary Material}
\author{\IEEEauthorblockN{Silvia Cascianelli$^1$, Vittorio Pippi$^1$, Martin Maarand$^2$, Marcella Cornia$^1$, Lorenzo Baraldi$^1$,\\Christopher Kermorvant$^2$, Rita Cucchiara$^1$}
\IEEEauthorblockA{$^1$University of Modena and Reggio Emilia, Modena, Italy \quad \quad $^2$TEKLIA, Paris, France \\
Email: $^1$\{name.surname\}@unimore.it, $^2$\{surname\}@teklia.com}
}

\begin{figure}[htb]
\twocolumn[{
\renewcommand\twocolumn[1][]{#1}%
\maketitle
\begin{center}
\begin{tabular}{cc}
\includegraphics[width=0.9\columnwidth]{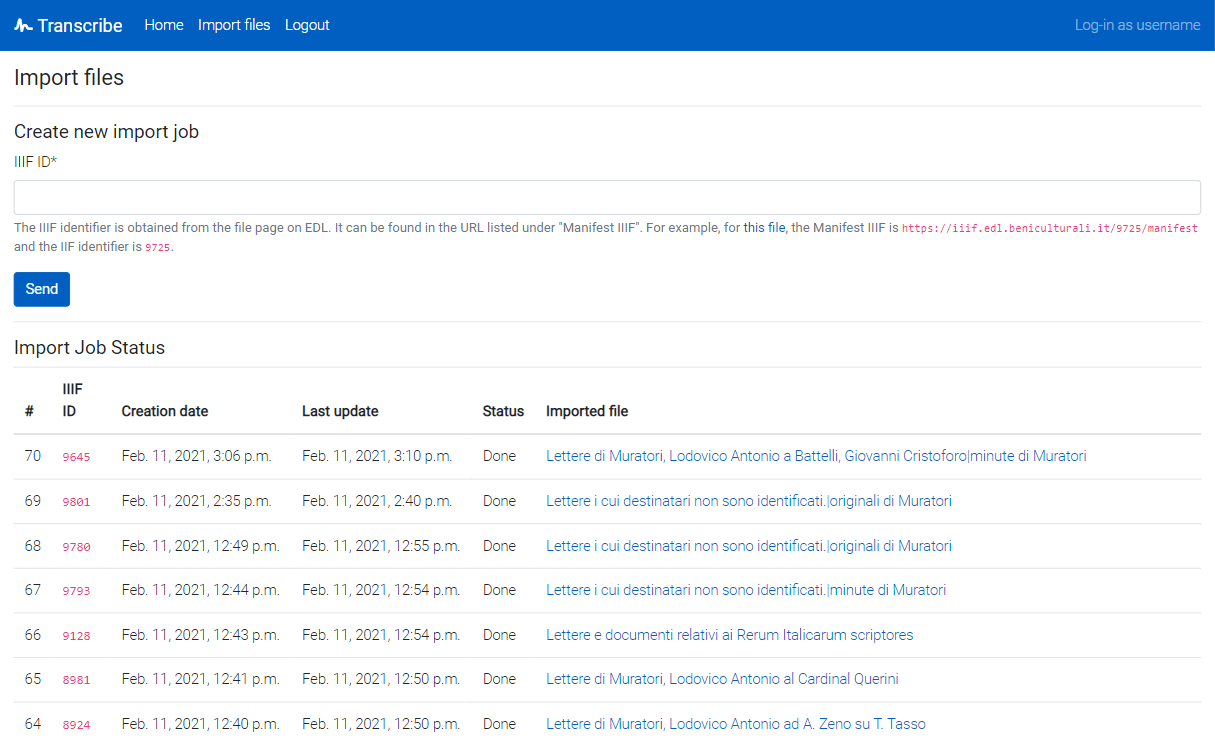} & 
\includegraphics[width=0.9\columnwidth]{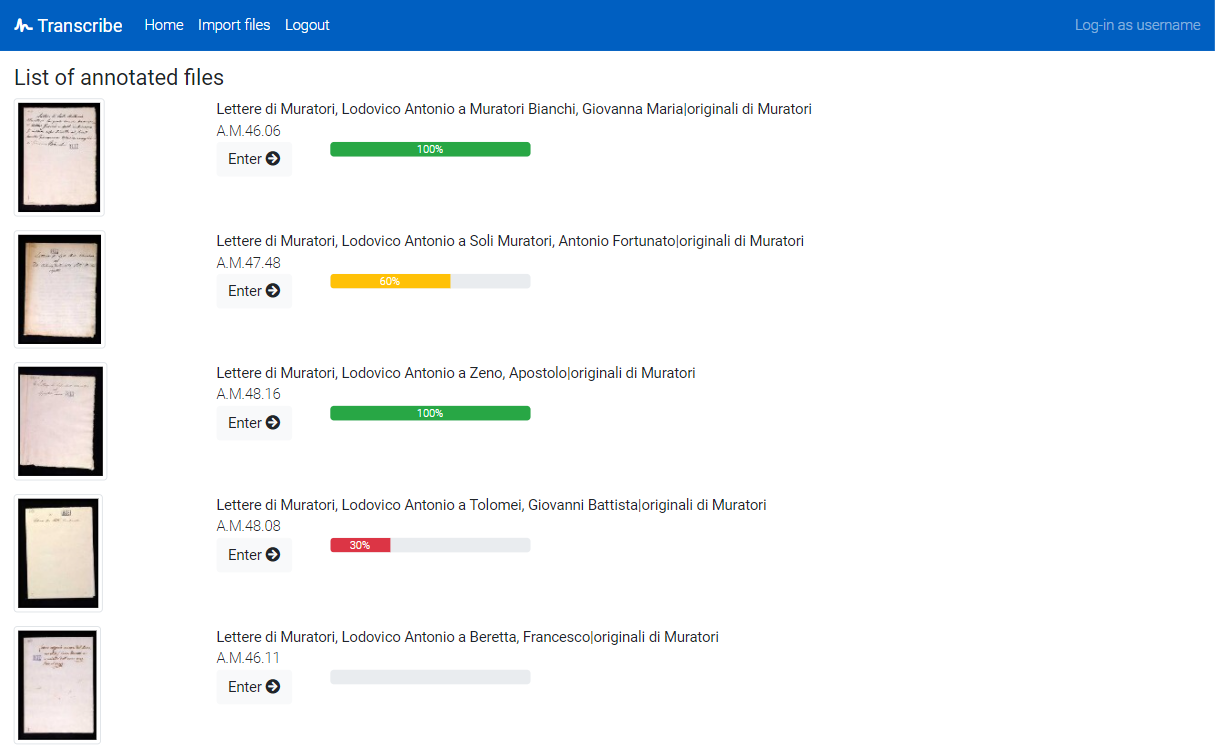} \\
\addlinespace[0.2cm]
\includegraphics[width=0.9\columnwidth]{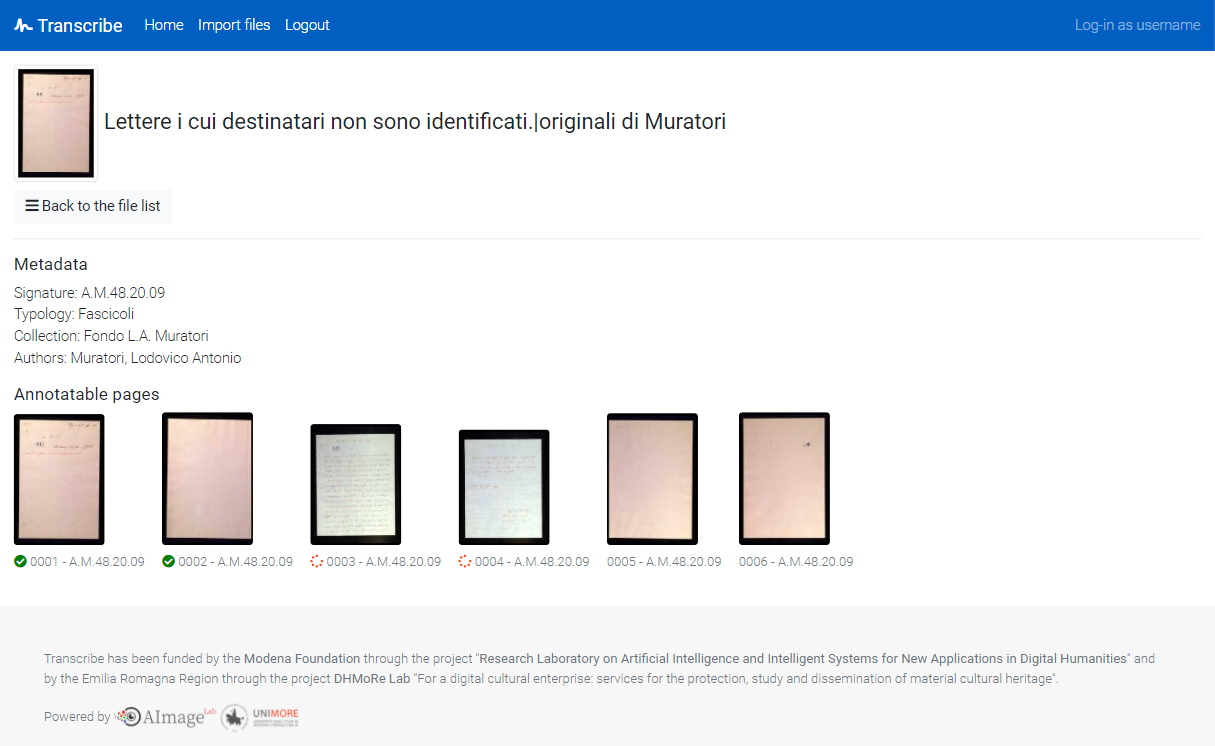} &
\includegraphics[width=0.9\columnwidth]{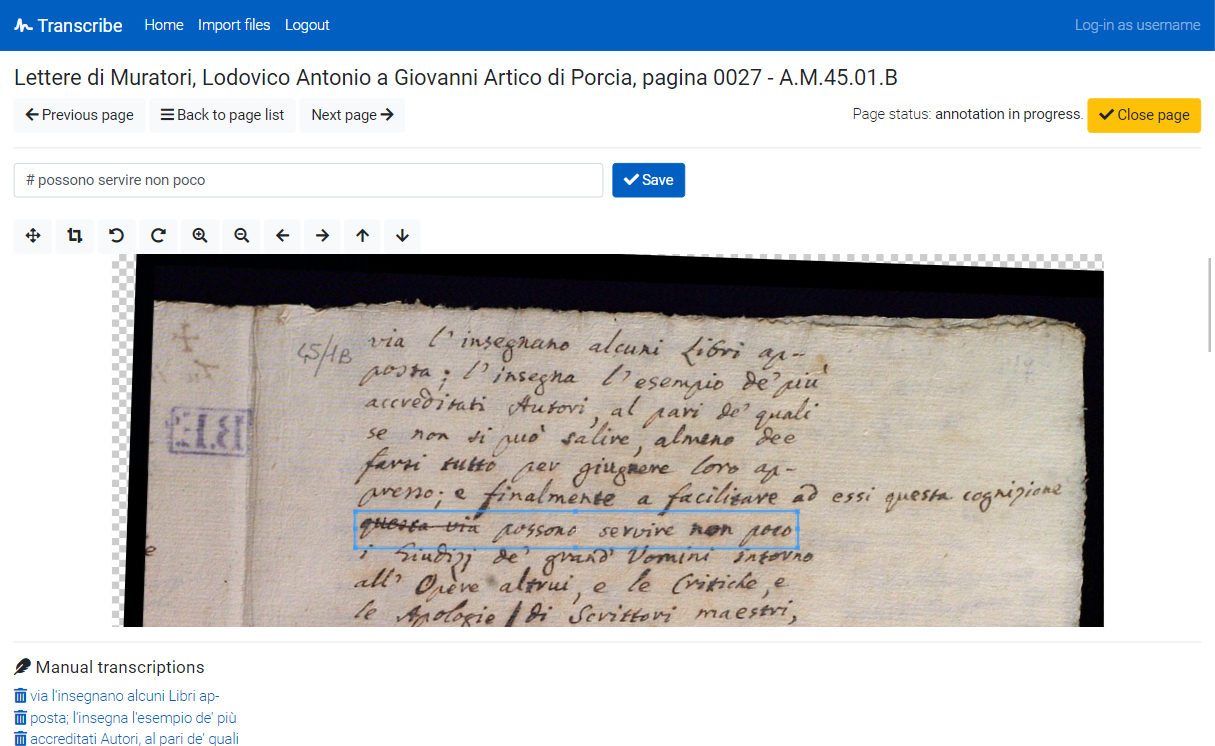} \\
\end{tabular}
\end{center}
\vspace{-0.3cm}
\caption{Screenshots of the annotation platform developed and used to annotate the LAM dataset.}
\label{fig:platform}
\vspace{1cm}
}]
\end{figure}

\vspace{-1.2cm}
\section{Details on the Annotation Platform}

In this section, we give further details on the features of the annotation platform developed to collect the LAM dataset.
Our tool allows the annotator to draw bounding boxes on each line, which can be eventually rotated, and to annotate each of these with a textual description. The interface further allows to easily browse documents in the collection and provides statistics on the annotation progress of each page.
The platform is linked to the digital library containing the documents of interest (the Estense Digital Library\footnote{\url{https://edl.beniculturali.it/}}, in the case of those used for the LAM dataset). From such collection, files of documents can be directly imported into the platform by specifying their IIIF identifier (see Fig.~\ref{fig:platform}, top left).

The list of imported files is visible on the main page of the platform, alongside progression bars indicating the portion of pages in the files that have already been fully annotated (see Fig.~\ref{fig:platform}, top right). This way, multiple users can collaborate on the annotation process and monitor their progress.
For each file, we visualize the main metadata and an overview of the contained pages. For each page, it is indicated the annotation status, which can be complete, in progress, or not yet started (Fig.~\ref{fig:platform}, bottom left). Again, this allows multiple users to collaborate on annotating pages in the same file. 

The core of the platform is the annotation interface (Fig.~\ref{fig:platform}, bottom right). To annotate the pages, the user can rotate the page to align the baseline of each line and then trace a bounding box around the line to be annotated. The transcription can be inserted in a dedicated text-box and saved. Previously inserted transcriptions are visible below the image of the page and can be edited after the first insertion (both the bounding box and the transcription). Once the page is fully annotated, it can be ``closed'' to make its annotations available for download. Note that, eventually, the pages can be re-opened for further editing.

The annotations of each file are exported as images of pages and XML files containing the line-level annotation. For each line, are indicated the width, the height, the orientation, and the coordinates of the top-left corner of its bounding box. 
This information is employed to extract lines images: overall, the dataset results in 25,823 JPEG files, which are on average 658$\pm$247 pixels wide and 53$\pm$16 pixels high.
Annotations are released as XML files that report the ground-truth transcription, the indication of the decade in which the page has been written, and the identifier of the user who inserted the transcription.

\begin{figure}[t]
\centering
\includegraphics[width=\columnwidth]{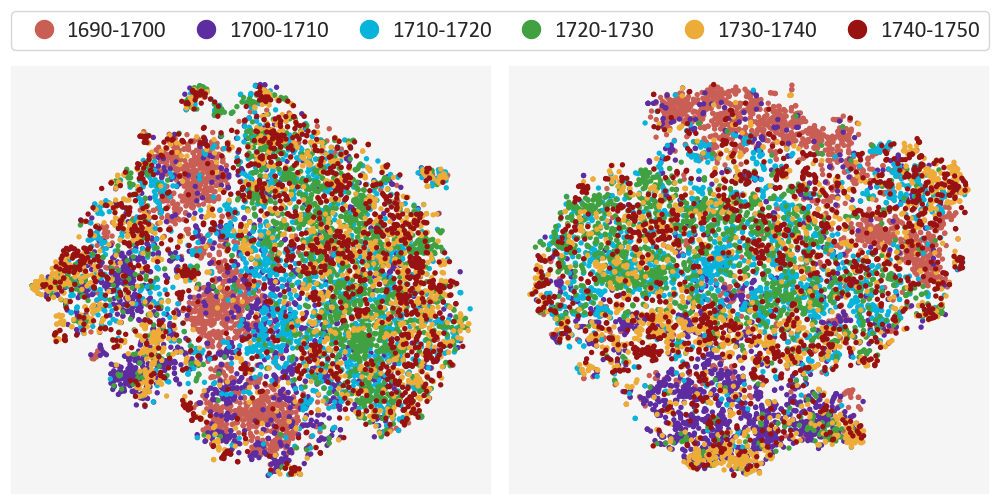} 
\caption{T-SNE plots of sample lines from the six time periods in the LAM dataset. The right plot is obtained with features from the convolutional part of the CRNN model, while the left plot is obtained with features from the convolutional part of the 1D-LSTM model. Best seen in color.}
\label{fig:tsnes}
\end{figure}

\section{Further Dataset Analisys}

\subsection{Word-level Analysis}
\begin{figure}[t]
\centering
\includegraphics[width=\columnwidth]{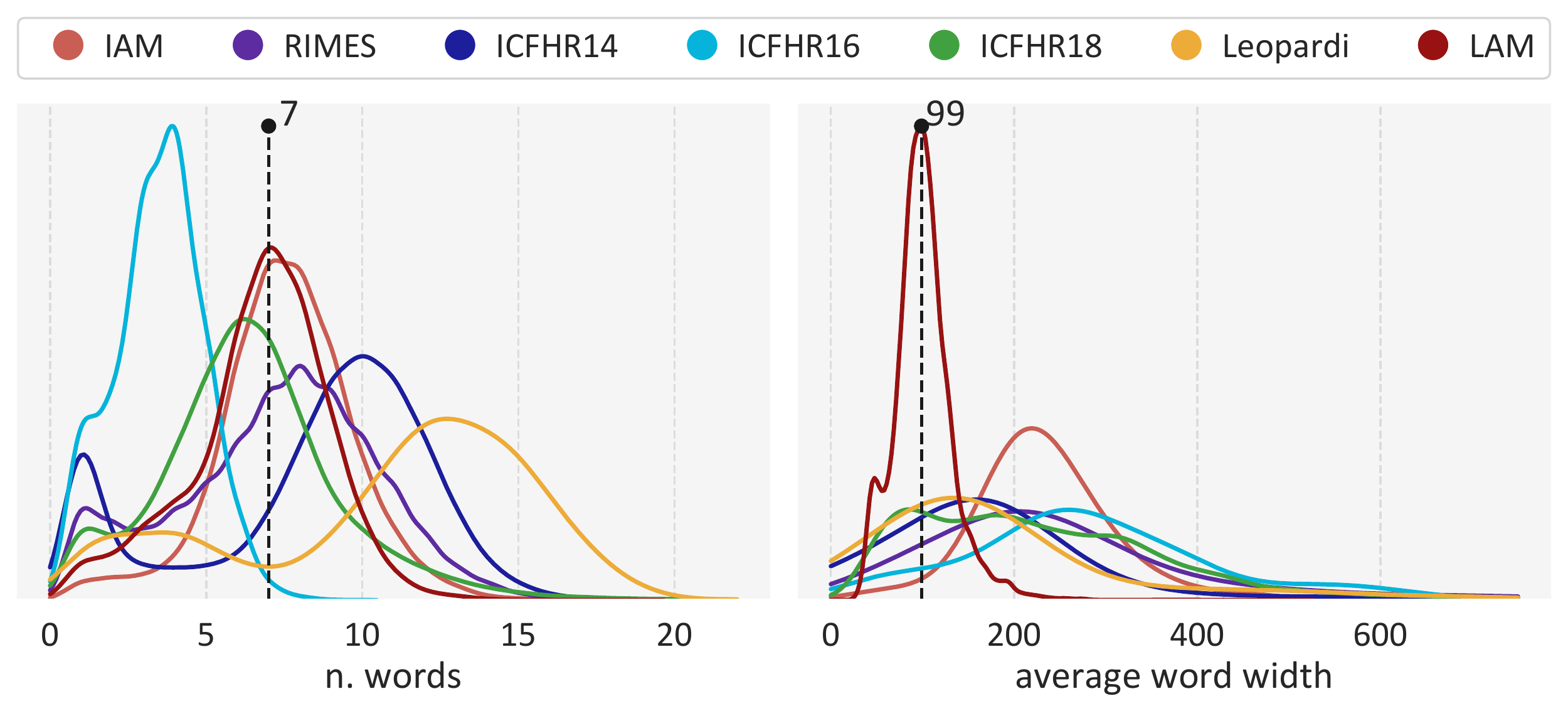} 
\caption{Number of words per line distribution in the LAM dataset compared to other popular benchmark datasets (left). Average words pixel width distribution in the LAM dataset compared to other popular benchmark datasets (right). Best seen in color.}
\label{fig:word_datasets_comparison}
\end{figure}

In Fig.~\ref{fig:word_datasets_comparison} we report word-level characteristics of the LAM dataset in comparison with other commonly used benchmark datasets, both modern and historical, and with a smaller historical dataset in Italian (the Leopardi dataset). Similar to what is observed in terms of number of characters per line and their average width, the LAM dataset exhibits more evident regularity when compared to other datasets. In fact, the majority of lines have 7 words (similar to IAM), whose length is bimodal-distributed.

\subsection{Date-based Setting}
The aim of the date-based setting is to allow exploring single-author HTR over long periods in the scenario in which samples are available for a specific time-span only. This is a challenging scenario despite targeting a single author since their handwriting can change over time, as in the case of the LAM dataset.  
This can be observed qualitatively from Fig.~\ref{fig:tsnes}, where we report a t-SNE analysis of a subset of text line images from the six decades. For this analysis, we use the feature maps from the last convolutional layer of the models presented in~\cite{shi2016end} and~\cite{puigcerver2017multidimensional}, averaged along the width. The models have been trained on a large synthetic dataset of 111,465 text lines in Italian in various fonts, as described in~\cite{cascianelli2021learning}. In both cases, some clusters can be observed, especially for the lines from the first and the second decade. Part of the lines from the other decades also form visible groups (\eg~the fifth decade). 
This visualization can help explain the performance of HTR models in the Date-based Setting. In fact, as reported in the main paper, the most challenging lines of the setting come from the first and second decades, whose features are indeed organized in clusters that are more evident than those of the other decades.

\subsection{Exemplar Pages from the LAM Dataset}
Some examples of pages from which the LAM dataset has been collected are reported in Fig.~\ref{fig:variability}. The pages have been grouped in decades as for the date-based setting. Differences in the handwriting style of the lines contained in the dataset can be appreciated, which makes the dataset as challenging as those containing text from multiple authors. Moreover, it can be noticed that the paper support is damaged (with scratches, creases, stains), and there are ink stains and bleeding.

\begin{table*}[t]
\centering
\small
\setlength{\tabcolsep}{.4em}
\caption{Results of OrigamiNet$_{\text{18}}$ on the decade-vs-decade setup (with training sets of different size) and then evaluated on the others.}
\label{tab:results_date}
\resizebox{0.75\textwidth}{!}{
\begin{tabular}{lc cc c cc c cc c cc c cc c cc c cc}
\toprule 
& & \multicolumn{2}{c}{\textbf{1690-1700}} & & \multicolumn{2}{c}{\textbf{1700-1710}} & & \multicolumn{2}{c}{\textbf{1710-1720}}  & & \multicolumn{2}{c}{\textbf{1720-1730}} & & \multicolumn{2}{c}{\textbf{1730-1740}} & & \multicolumn{2}{c}{\textbf{1740-1750}} \\
\cmidrule{3-4} \cmidrule{6-7} \cmidrule{9-10} \cmidrule{12-13} \cmidrule{15-16} \cmidrule{18-19} 
\textbf{Train} & & \textbf{CER} & \textbf{WER} & & \textbf{CER} & \textbf{WER} & & \textbf{CER} & \textbf{WER} & & \textbf{CER} & \textbf{WER} & & \textbf{CER} & \textbf{WER} & & \textbf{CER} & \textbf{WER} \\
\midrule

\hspace{0.4cm}\textbf{1690-1700}    & &  -   &  -   & & 3.9 & 14.7 & & 3.8   & 14.0   & & 4.3   & 15.5   & & 6.9   & 23.2  & & 6.4   & 23.7  \\
\hspace{0.4cm}\textbf{1700-1710}    & & 5.8  & 23.3 & &  -   &  -  & & 3.6   & 13.3   & & 3.9   & 14.4   & & 6.6   & 23.0  & & 6.2   & 23.2  \\
\hspace{0.4cm}\textbf{1710-1720}    & & 8.9  & 33.4 & & 5.2 & 19.5 & &  -    &  -     & & 4.0   & 15.2   & & 7.2   & 24.2  & & 7.4   & 26.2  \\
\hspace{0.4cm}\textbf{1720-1730}    & & 7.7  & 29.6 & & 4.5 & 17.0 & & 3.3   & 12.1   & &   -   &  -     & & 5.6   & 19.3  & & 6.2   & 21.9  \\
\hspace{0.4cm}\textbf{1730-1740}    & & 7.3  & 28.6 & & 4.8 & 18.8 & & 3.4   & 13.1   & & 3.2   & 12.7   & &   -   &  -    & & 4.6   & 16.7  \\
\hspace{0.4cm}\textbf{1740-1750}    & & 6.9  & 26.7 & & 4.5 & 17.3 & & 3.3   & 13.0   & & 3.4   & 12.9   & & 4.2   & 15.0  & &   -   &  -    \\
\bottomrule
\end{tabular}
}
\end{table*}

\begin{table*}[t]
\centering
\small
\setlength{\tabcolsep}{.4em}
\caption{Results of OrigamiNet$_{\text{18}}$ on the decade-vs-decade setup (with training sets of the same size) and then evaluated on the others.}
\label{tab:results_date_balanced}
\resizebox{0.75\textwidth}{!}{
\begin{tabular}{lc cc c cc c cc c cc c cc c cc c cc}
\toprule 
& & \multicolumn{2}{c}{\textbf{1690-1700}} & & \multicolumn{2}{c}{\textbf{1700-1710}} & & \multicolumn{2}{c}{\textbf{1710-1720}}  & & \multicolumn{2}{c}{\textbf{1720-1730}} & & \multicolumn{2}{c}{\textbf{1730-1740}} & & \multicolumn{2}{c}{\textbf{1740-1750}} \\
\cmidrule{3-4} \cmidrule{6-7} \cmidrule{9-10} \cmidrule{12-13} \cmidrule{15-16} \cmidrule{18-19} 
\textbf{Train} & & \textbf{CER} & \textbf{WER} & & \textbf{CER} & \textbf{WER} & & \textbf{CER} & \textbf{WER} & & \textbf{CER} & \textbf{WER} & & \textbf{CER} & \textbf{WER} & & \textbf{CER} & \textbf{WER} \\
\midrule
\hspace{0.4cm}\textbf{1690-1700}    & &  -   &  -   & & 5.7 & 21.0 & & 6.5   & 23.5   & & 7.1   & 24.3   & & 10.3   & 33.8  & & 9.5   & 33.7  \\
\hspace{0.4cm}\textbf{1700-1710}    & & 7.0  & 27.5 & &  -   &  -  & & 4.4   & 16.4   & & 5.0   & 18.2   & & 7.9   & 26.9  & & 7.6   & 27.4  \\
\hspace{0.4cm}\textbf{1710-1720}    & & 8.3  & 31.9 & & 5.1 & 19.3 & &  -    &  -     & & 4.1   & 15.3   & & 7.2   & 24.6  & & 7.2   & 25.7  \\
\hspace{0.4cm}\textbf{1720-1730}    & & 9.1  & 34.2 & & 5.9 & 21.4 & & 4.5   & 16.0   & &  -    &  -     & & 6.8   & 22.9  & & 7.6   & 25.8  \\
\hspace{0.4cm}\textbf{1730-1740}    & & 8.9  & 33.2 & & 6.2 & 23.2 & & 4.7   & 17.5   & & 4.2   & 16.1   & &  -    &  -    & & 5.4   & 19.4  \\
\hspace{0.4cm}\textbf{1740-1750}    & & 8.5  & 31.9 & & 5.6 & 21.1 & & 4.9   & 18.4   & & 4.7   & 17.0   & & 5.3   & 18.5  & &  -    &  -     \\

\bottomrule
\end{tabular}
}
\end{table*}

\begin{figure*}[t]
\centering
\footnotesize
\setlength{\tabcolsep}{.7em}
\resizebox{\linewidth}{!}{
\begin{tabular}{llcll}
\multicolumn{2}{c}{\includegraphics[width=0.45\linewidth]{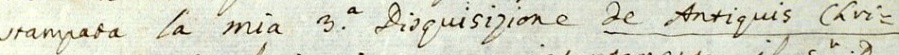}} & & \multicolumn{2}{c}{\includegraphics[width=0.45\linewidth]{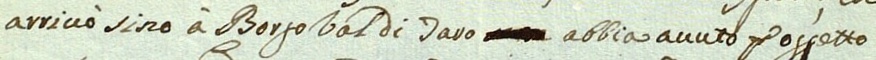}} \\
\textbf{Ground Truth}         & stampata la mia 3.a Disquisizione de Antiquis Chri= & & 
\textbf{Ground Truth}         & arriv\`o sino \`a Borho Tol di Taro \# abbia avuto p oggetto \\
\textbf{OrigamiNet$_{\text{24}}$} & stampata la mia 3.a Disquisizione de Antiquis Chri= & & 
\textbf{OrigamiNet$_{\text{24}}$} & arriv\`o sino a Borgolbal di Daro \# abbia avuto Doggetto \\
\\
\multicolumn{2}{c}{\includegraphics[width=0.45\linewidth]{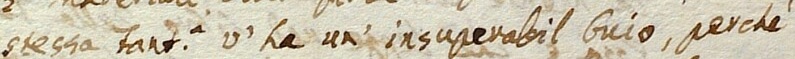}} & & \multicolumn{2}{c}{\includegraphics[width=0.45\linewidth]{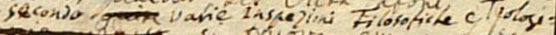}} \\
\textbf{Ground Truth}         & stessa Fant.a v'ha un' insuperabil buio, perché & & 
\textbf{Ground Truth}         & secondo \# varie inspezioni Filofiche e Teologi= \\
\textbf{OrigamiNet$_{\text{24}}$} & stessa Fant.a v'ha un' insuperabil buio, perché & & 
\textbf{OrigamiNet$_{\text{24}}$} & secondo \# varie inspazioni Filotofiche e l'olorp. \\
\\
\multicolumn{2}{c}{\includegraphics[width=0.45\linewidth]{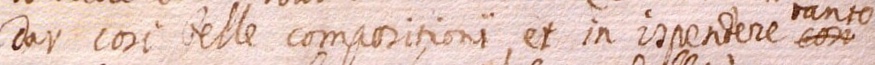}} & & \multicolumn{2}{c}{\includegraphics[width=0.45\linewidth]{images/qualitatives/074_3942_01.jpg}} \\
\textbf{Ground Truth}         & dar così belle composizioni, et in ispendere tanto. & & 
\textbf{Ground Truth}         & da altri \# per benefizio delle Lettere, \\
\textbf{OrigamiNet$_{\text{24}}$} & dar così belle composizioni, et in ispendere tanto & & 
\textbf{OrigamiNet$_{\text{24}}$} & da altri MS\#.i, per benefizio delle Lettere \\
\\
\multicolumn{2}{c}{\includegraphics[width=0.45\linewidth]{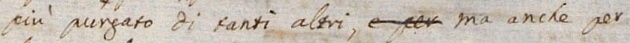}} & & \multicolumn{2}{c}{\includegraphics[width=0.45\linewidth]{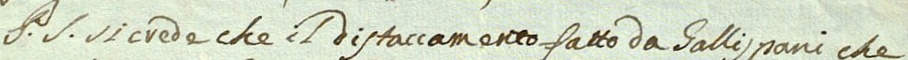}} \\
\textbf{Ground Truth}         & più purgato di tanti altri, \# ma anche per & & 
\textbf{Ground Truth}         & P.S. si crede che il distaccamento fatto da Galljpani, che \\
\textbf{OrigamiNet$_{\text{24}}$} & più purgato di tanti altri, \# ma anche per & & 
\textbf{OrigamiNet$_{\text{24}}$} & S. S. Si crede che il distaccamanto Datto da Sallizpani che \\

\end{tabular}
}
\caption{Qualitative results on example lines from the LAM dataset of the best performing model on the basic split. Examples on the right are among the most challenging (CER$>$15\%).}
\label{fig:qualitatives}
\end{figure*}

\begin{table}[t]
\centering
\small
\setlength{\tabcolsep}{.3em}
\caption{Statistics of the balanced decade-vs-decade setup of the date-based setting. The charset size is calculated on the Training and Validation splits.}
\label{tab:splittings_balanced}
\resizebox{0.85\columnwidth}{!}{%
\begin{tabular}{lc ccccc}
\toprule 
& & \textbf{Total} & \textbf{Training} & \textbf{Validation} & \textbf{Test} & \textbf{Charset} \\
\midrule
\textit{1690-1700} & & 21,066 & 1,755 & 195 & 19,116 & 78\\
\textit{1700-1710} & & 23,857 & 1,755 & 195 & 21,907 & 78\\
\textit{1710-1720} & & 25,183 & 1,755 & 195 & 23,233 & 79\\
\textit{1720-1730} & & 21,091 & 1,755 & 195 & 19,141 & 78\\
\textit{1730-1740} & & 23,275 & 1,755 & 195 & 21,325 & 79\\
\textit{1740-1750} & & 23,143 & 1,755 & 195 & 21,193 & 77\\
\bottomrule
\end{tabular}
}
\vspace{-.5cm}
\end{table}

\section{Additional Results}

\subsection{Additional Results on the Date-based Setting}
In Table~\ref{tab:results_date} and Table~\ref{tab:results_date_balanced}, we report further details on the decade-vs-decade analysis described in the main paper. In particular, we include WER scores, which are in line with the CER scores and thus remark the characteristics and the challenges that can be explored within the date-based setting. Moreover, in Table~\ref{tab:splittings_balanced} we report the characteristics of the balanced variant of the decade-vs-decade setup of the date-based setting.

\subsection{Qualitative Results}
As a further analysis, we examine the test lines that can be more challenging for an HTR system, to give some insights on the source of such challenges. To this end, we select those lines where the best-performing model, OrigamiNet$_{\text{24}}$, obtained CER$\geq$15\%. Some examples are reported in Fig~\ref{fig:qualitatives}. It can be observed that the handwriting style makes uppercase letters more difficult to distinguish. Another issue is represented by those symbols that cannot be represented in Unicode and thus have been annotated as \texttt{\#}'s. The models struggle to learn to skip those symbols and transcribe them with combinations of Unicode characters. From the exemplar images, it can also be appreciated the differences in the handwriting, which makes the LAM dataset challenging despite the fact that it contains text written by a single author.

\begin{figure*}[t]
\centering
\footnotesize
\setlength{\tabcolsep}{.3em}
\resizebox{0.85\linewidth}{!}{
\begin{tabular}{cccccc}
\rotatebox{90}{\parbox[t]{1.3in}{\hspace*{\fill}\textbf{1690-1700}\hspace*{\fill}}} & 
\includegraphics[width=0.15\linewidth]{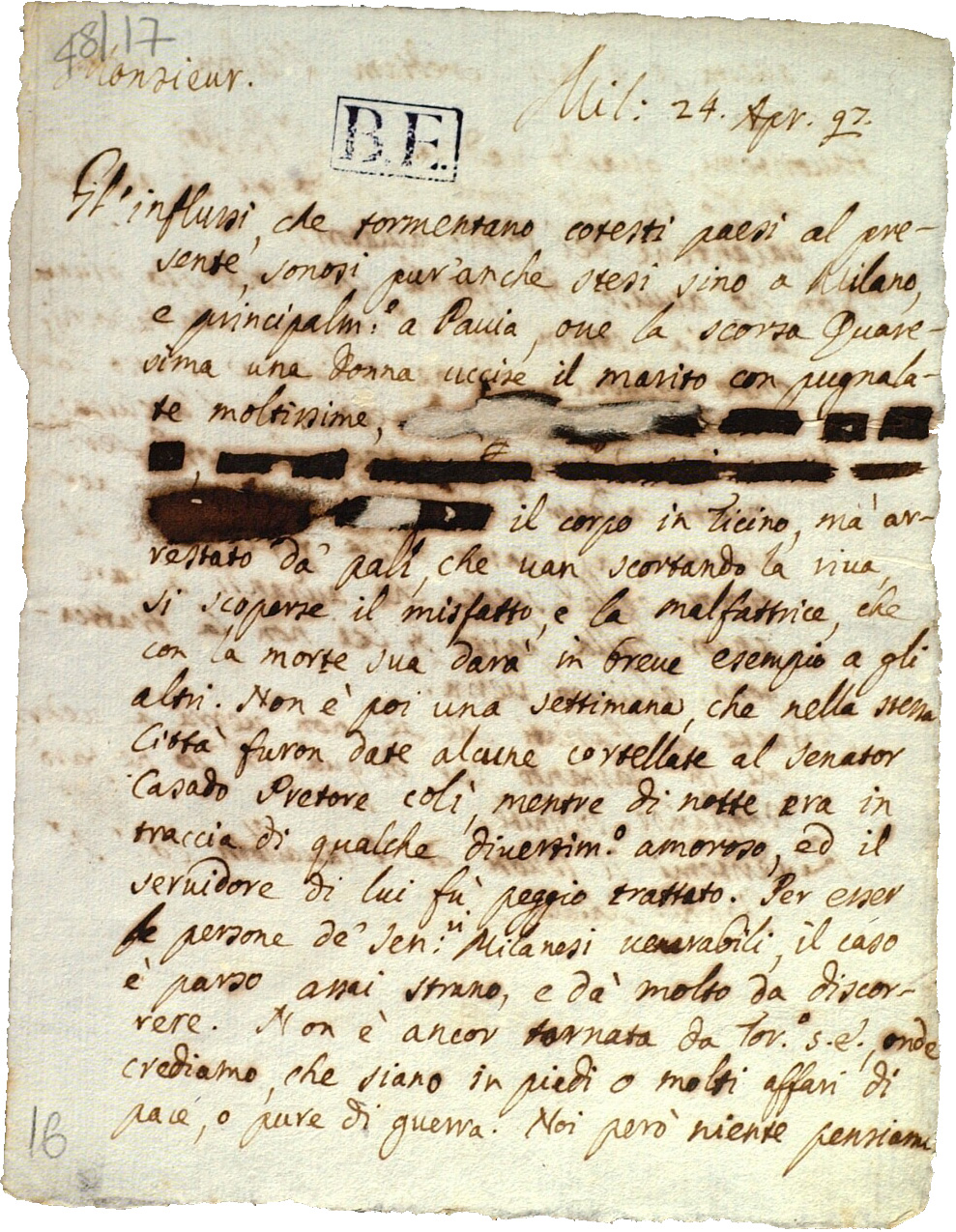} &
\includegraphics[width=0.15\linewidth]{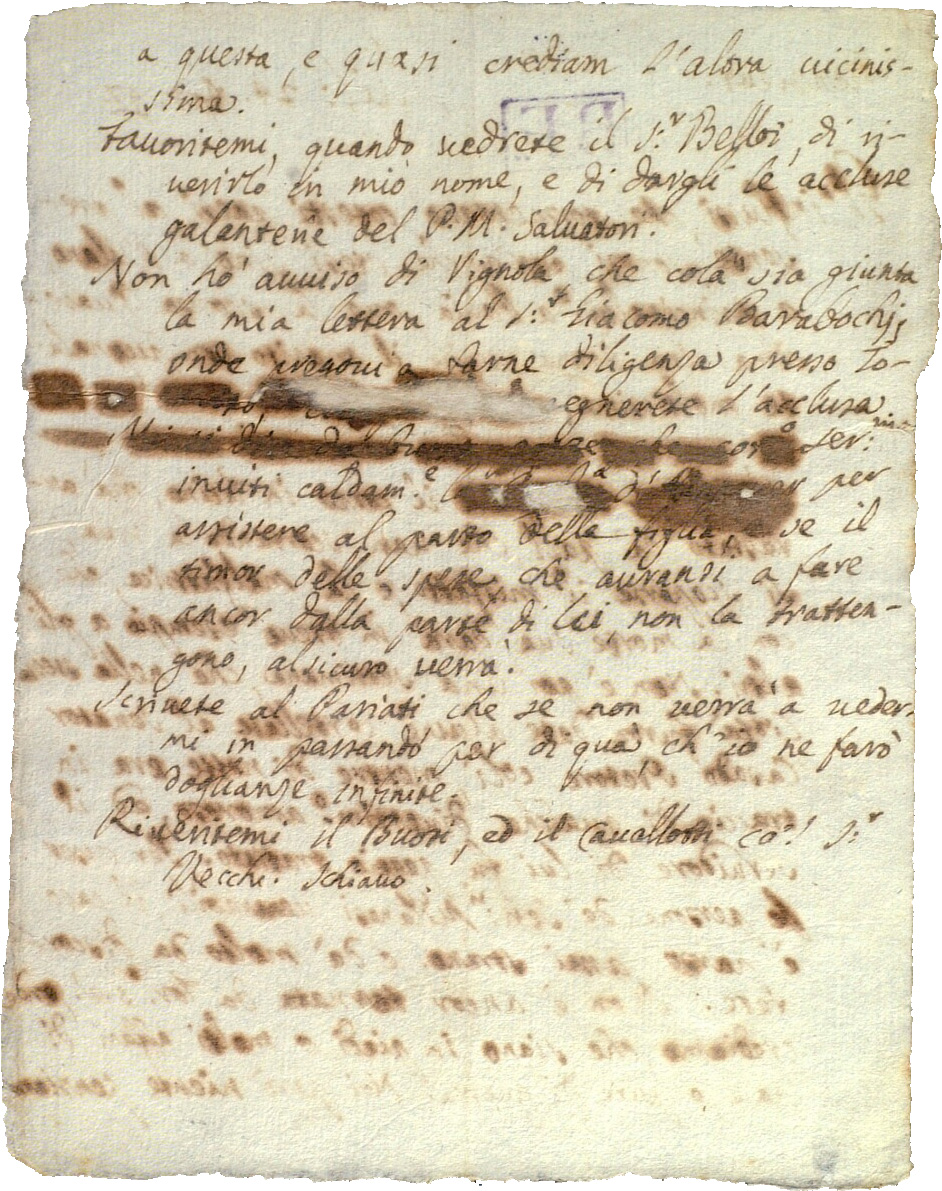} &
\includegraphics[width=0.15\linewidth]{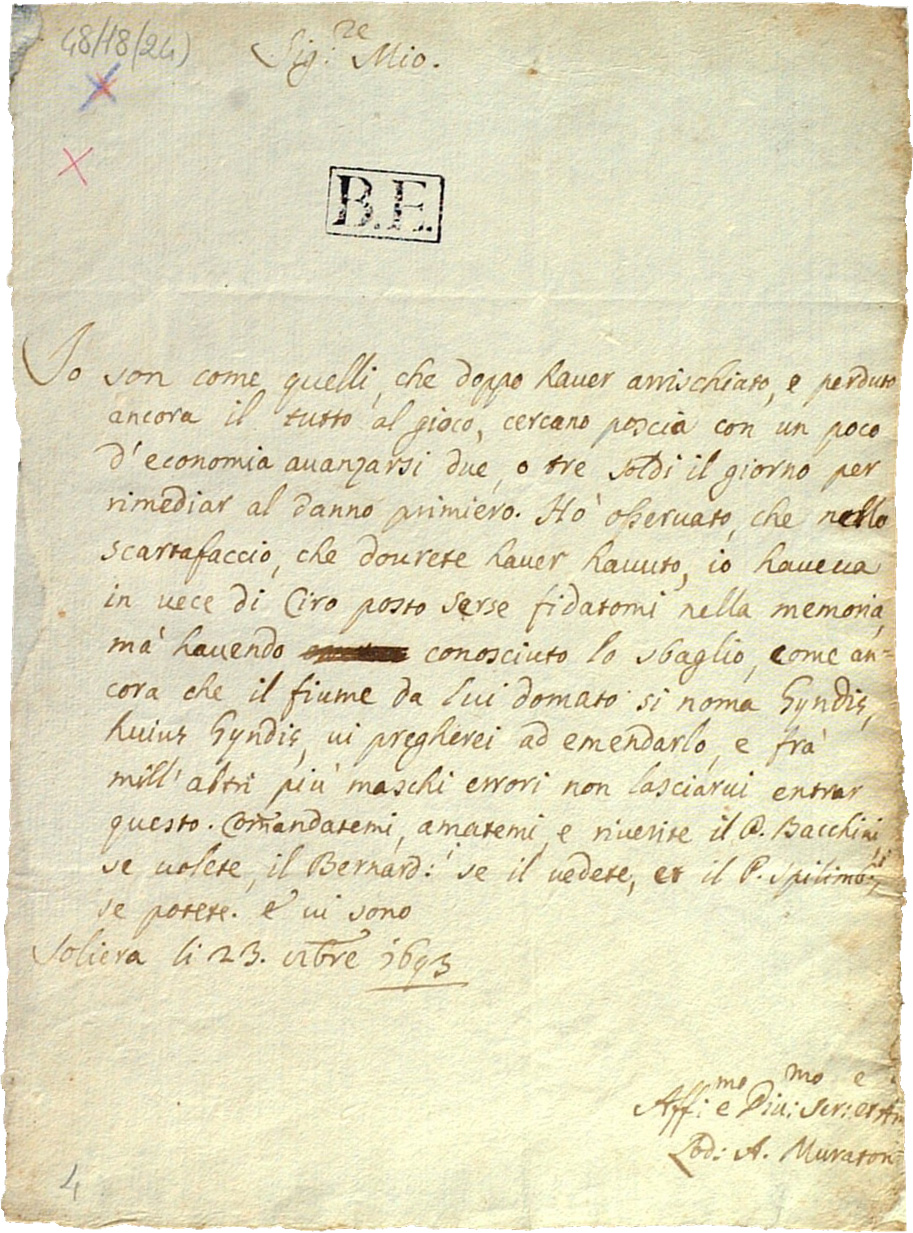} &
\includegraphics[width=0.15\linewidth]{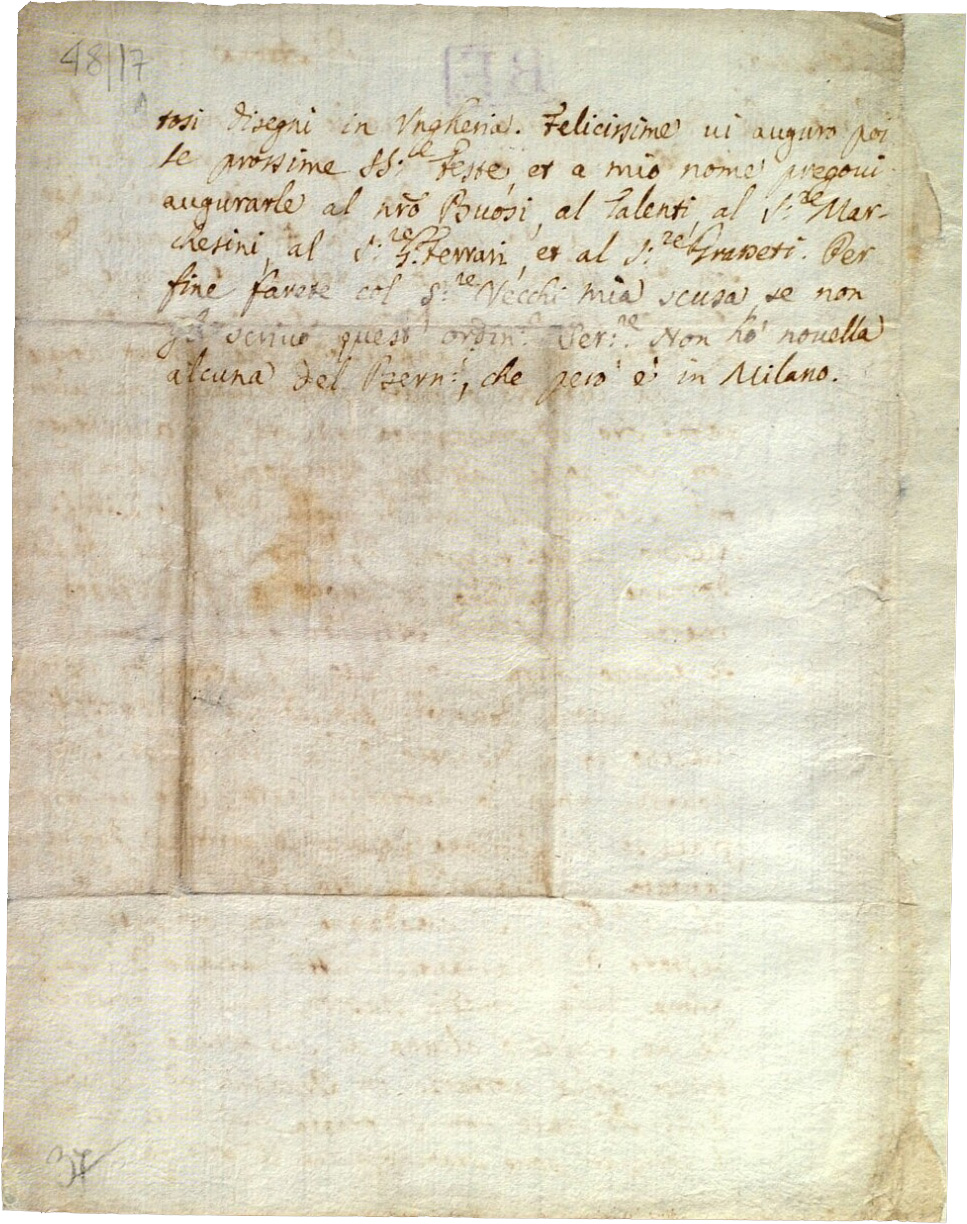} &
\includegraphics[width=0.15\linewidth]{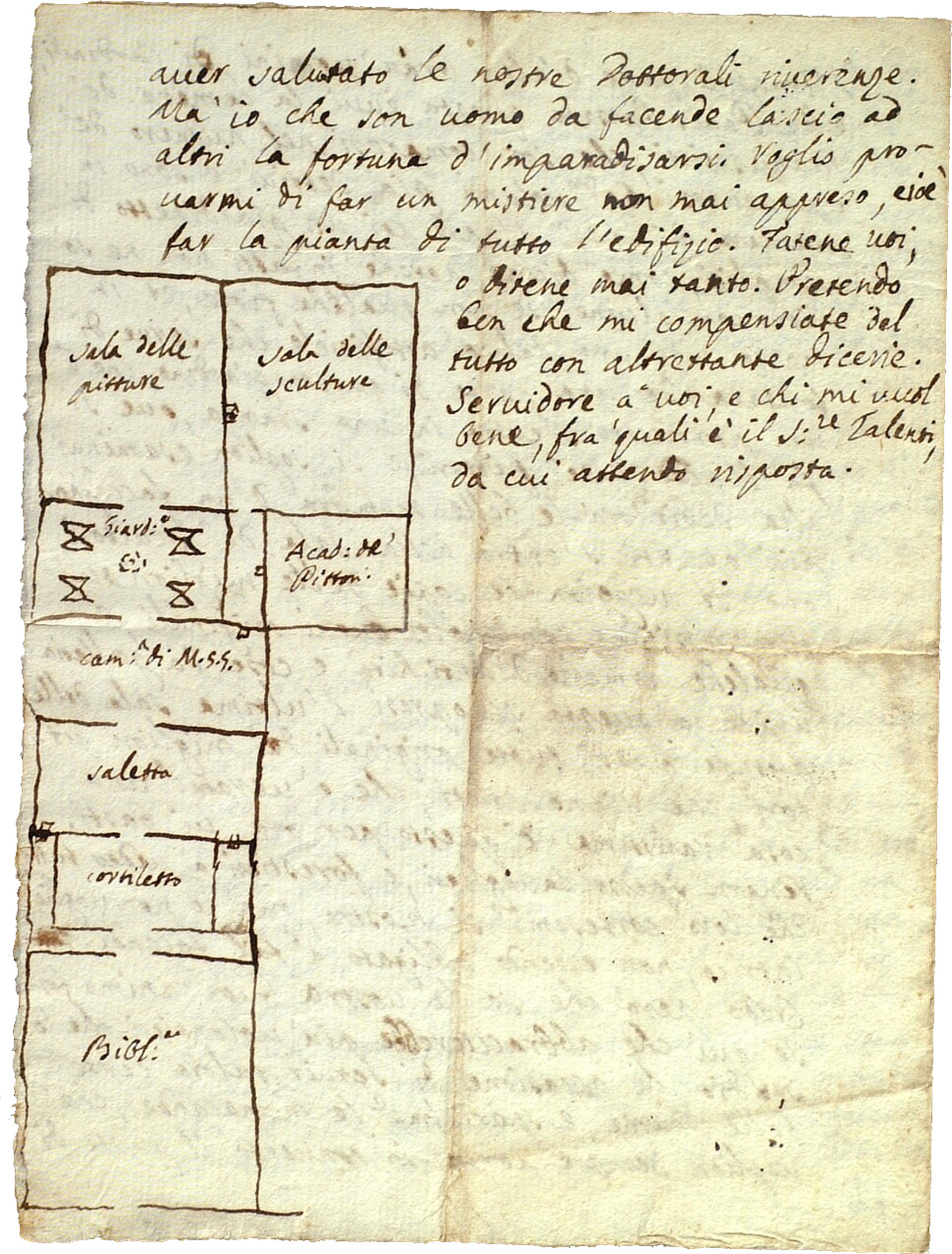} \\
\rotatebox{90}{\parbox[t]{1.4in}{\hspace*{\fill}\textbf{1700-1710}\hspace*{\fill}}}  & 
\includegraphics[width=0.15\linewidth]{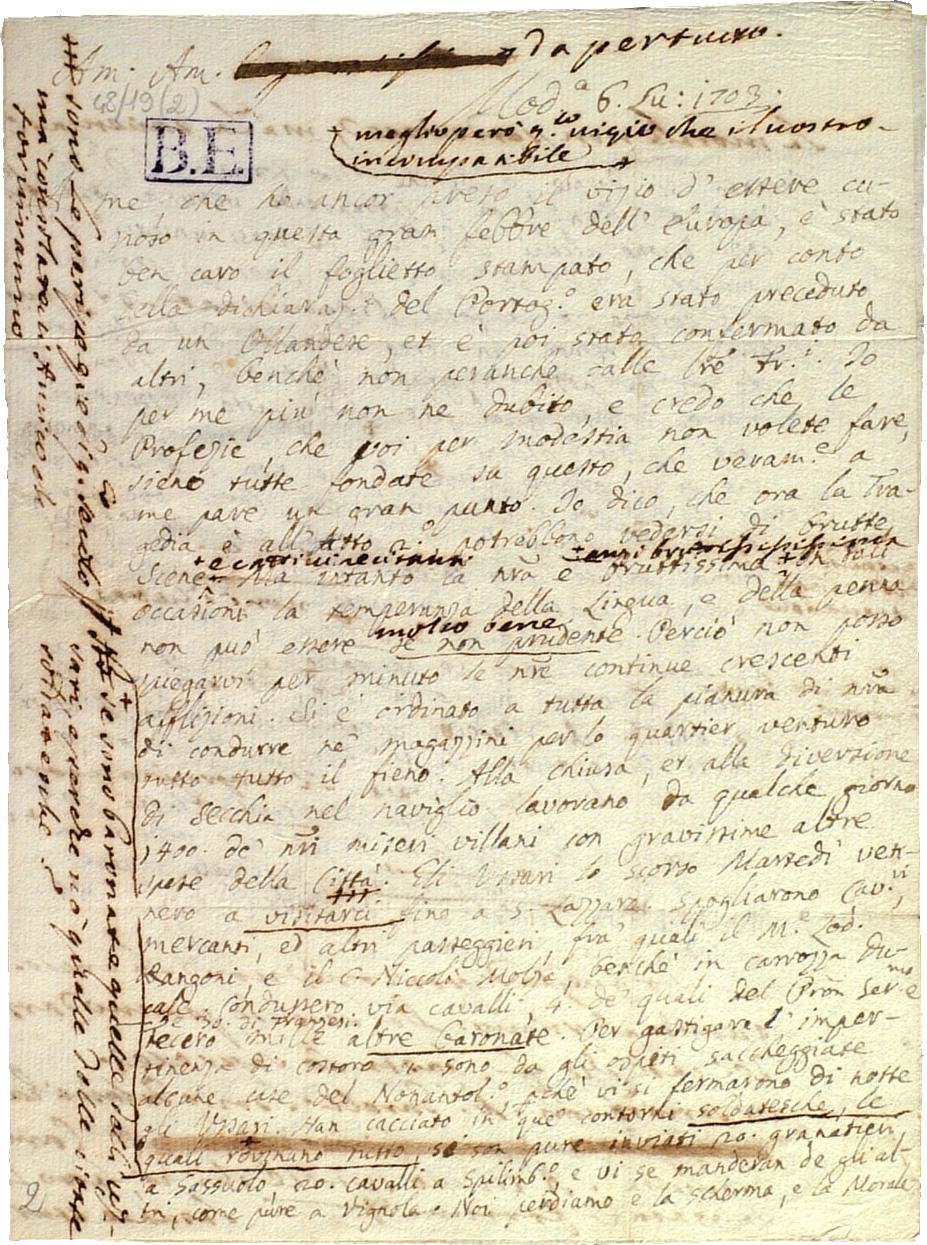} &
\includegraphics[width=0.15\linewidth]{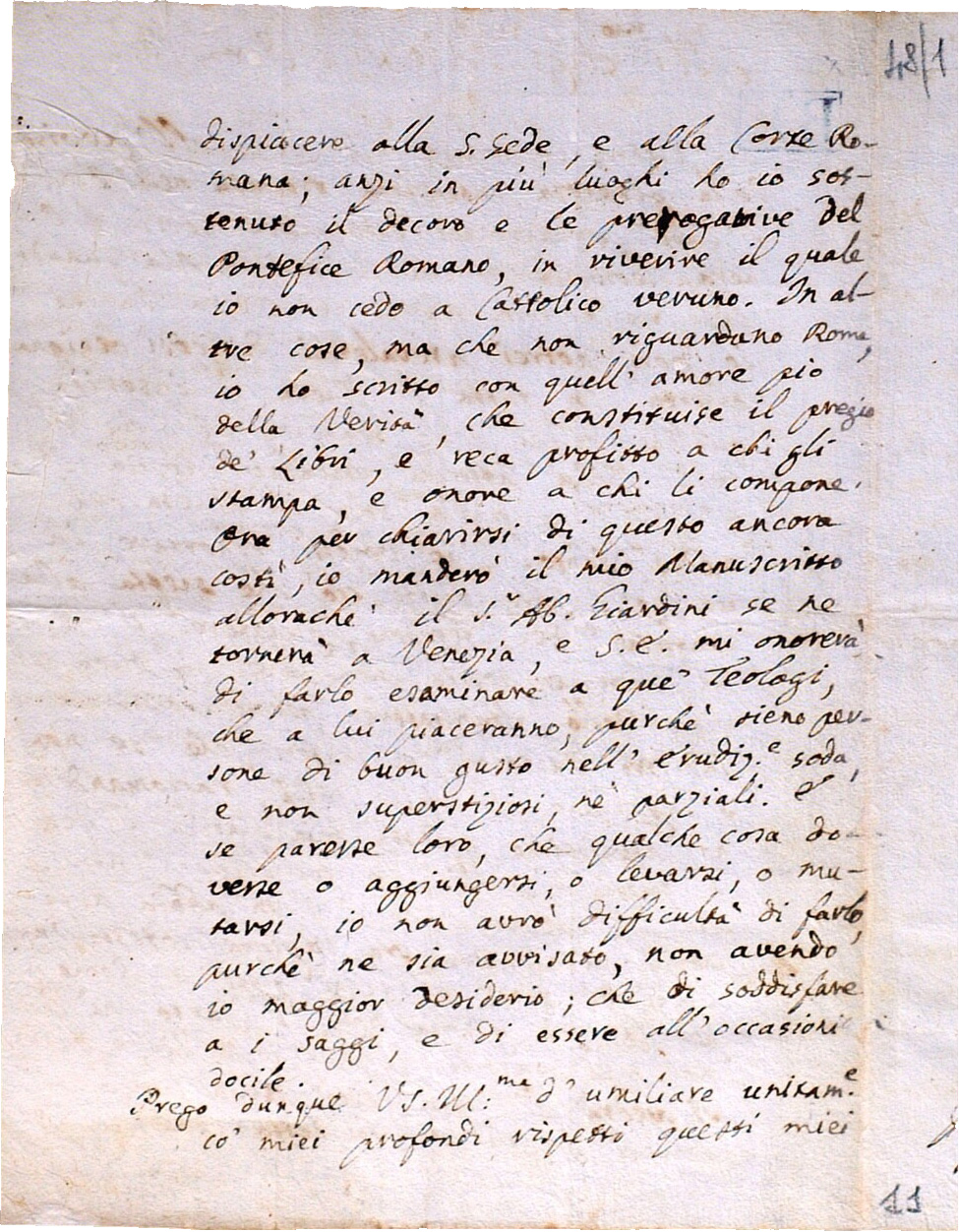} &
\includegraphics[width=0.15\linewidth]{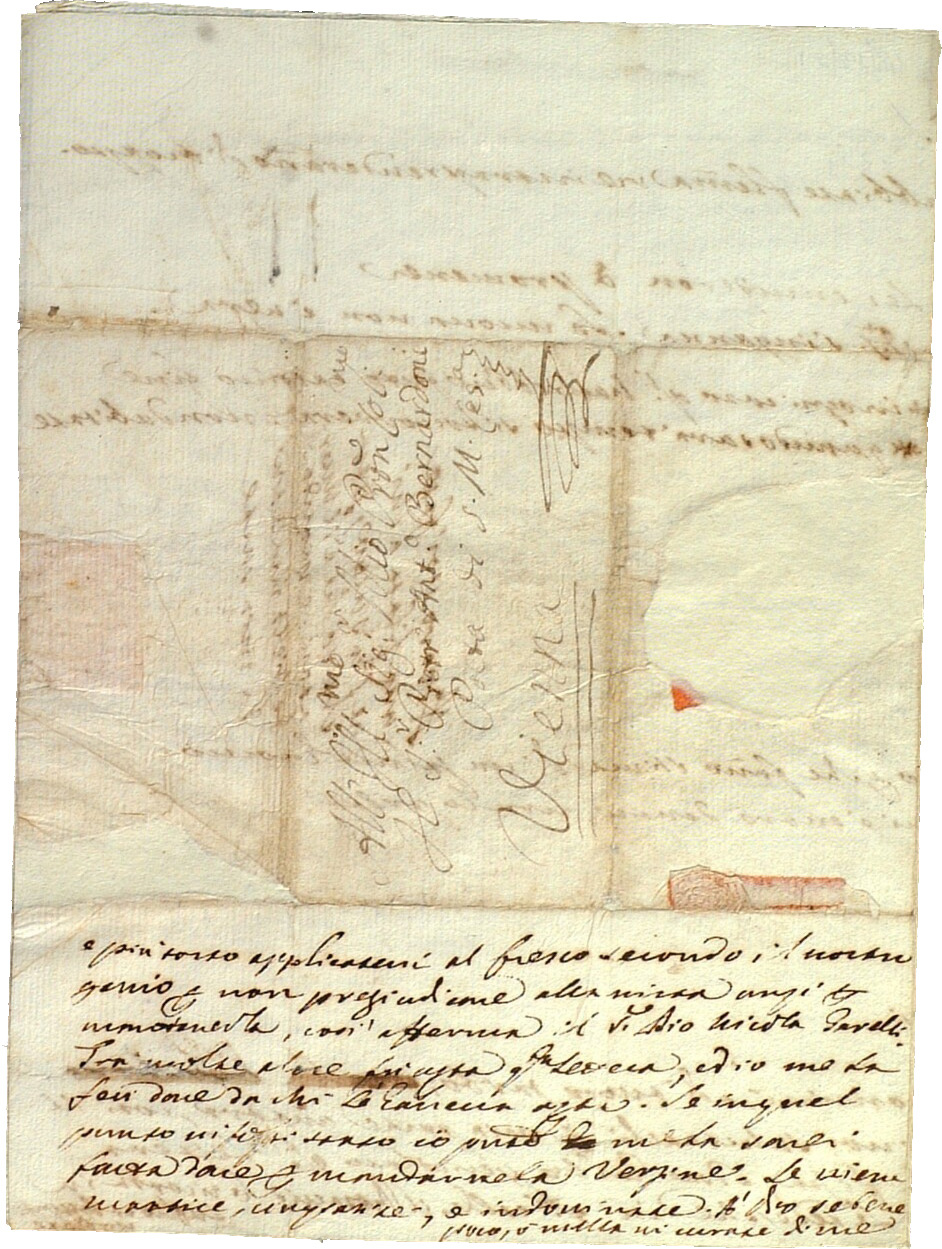} &
\includegraphics[width=0.15\linewidth]{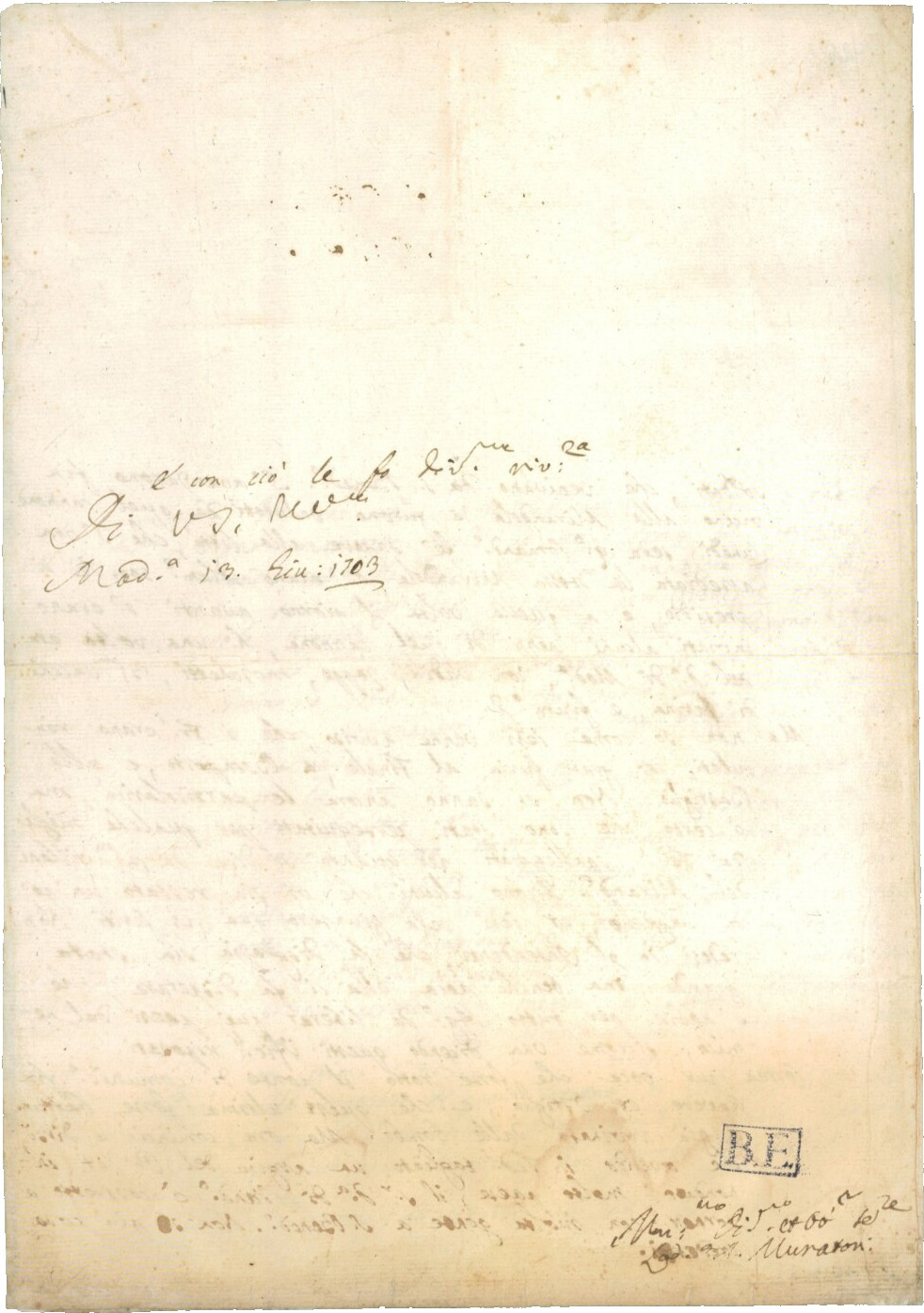} &
\includegraphics[width=0.15\linewidth]{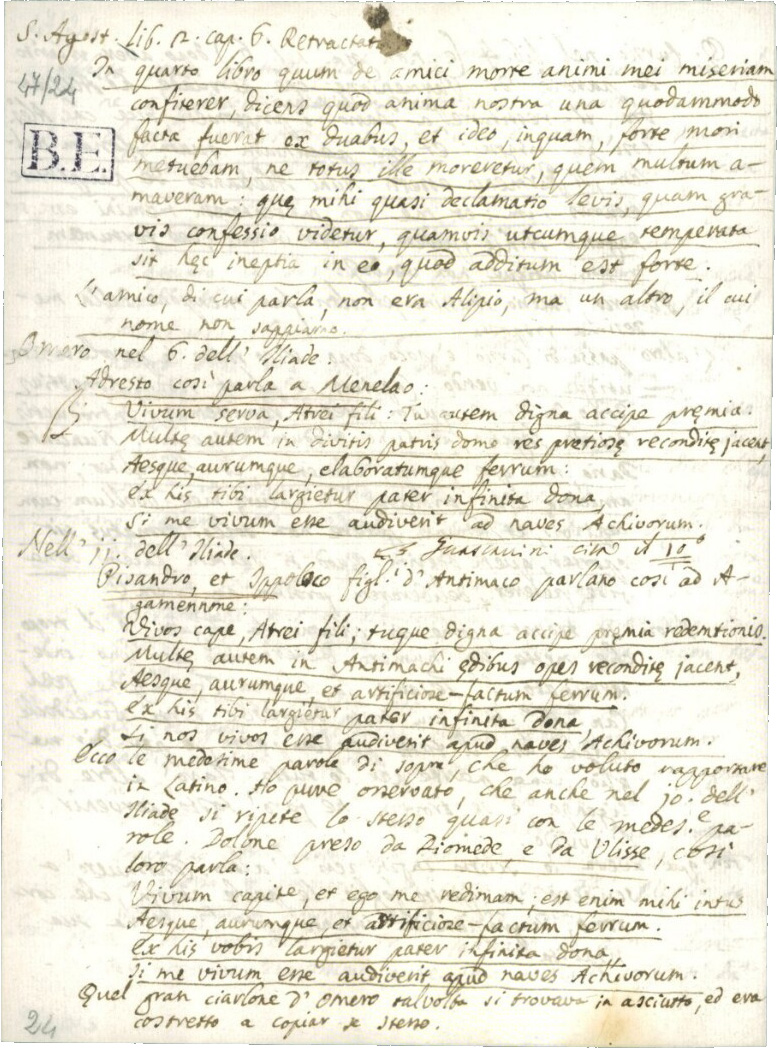} \\
\rotatebox{90}{\parbox[t]{1.35in}{\hspace*{\fill}\textbf{1710-1720}\hspace*{\fill}}}  & 
\includegraphics[width=0.15\linewidth]{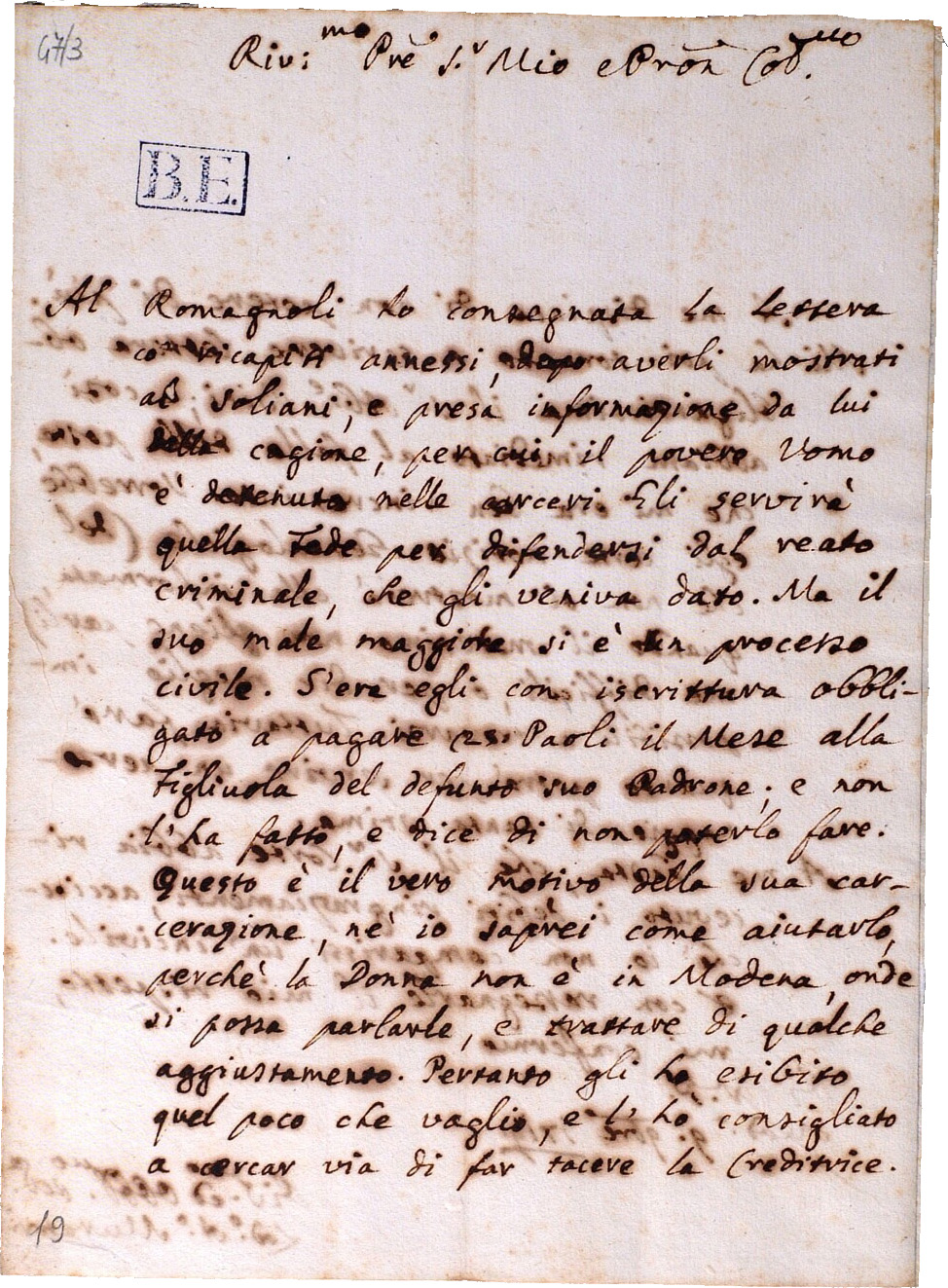} &
\includegraphics[width=0.15\linewidth]{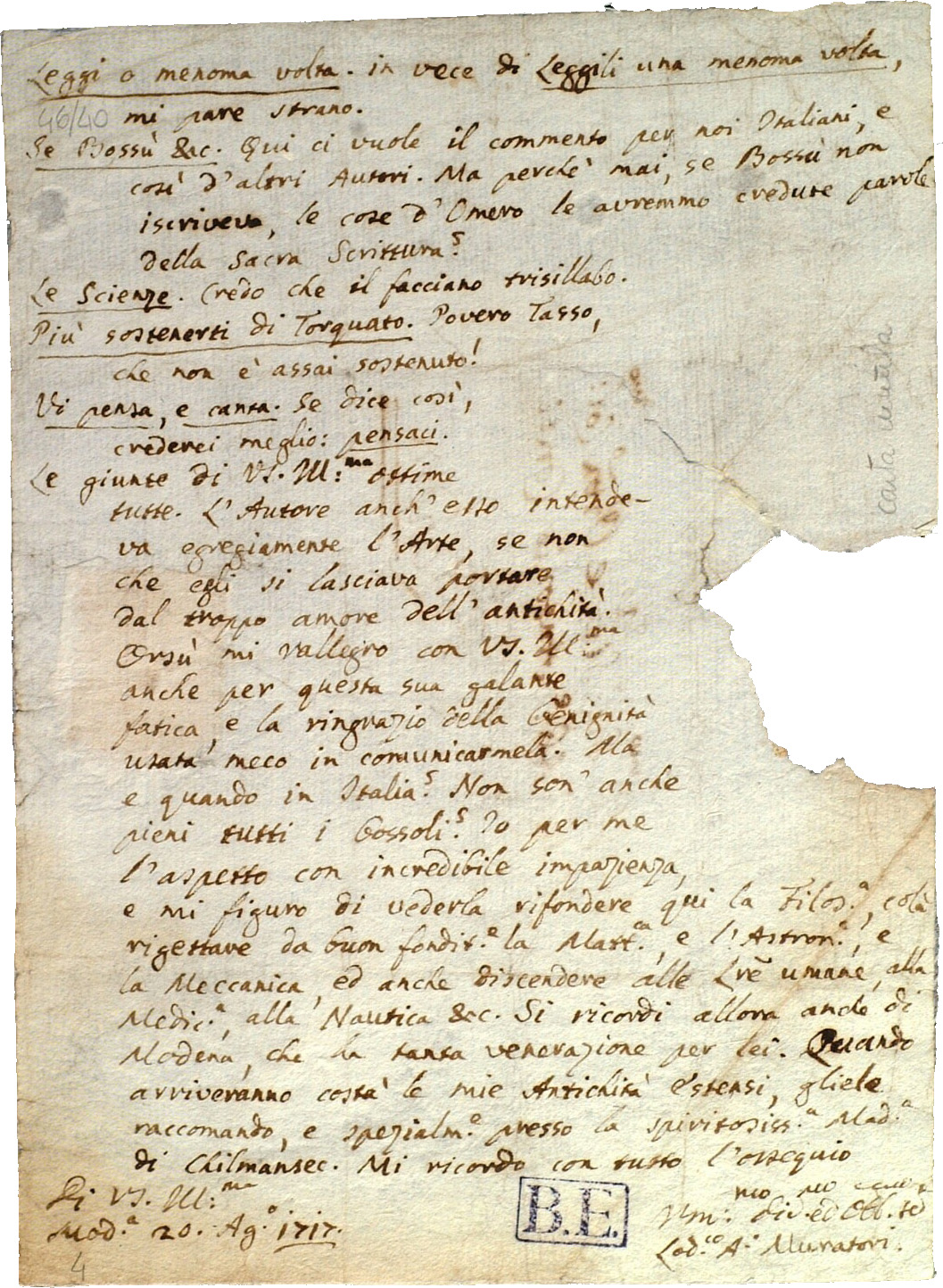} & 
\includegraphics[width=0.15\linewidth]{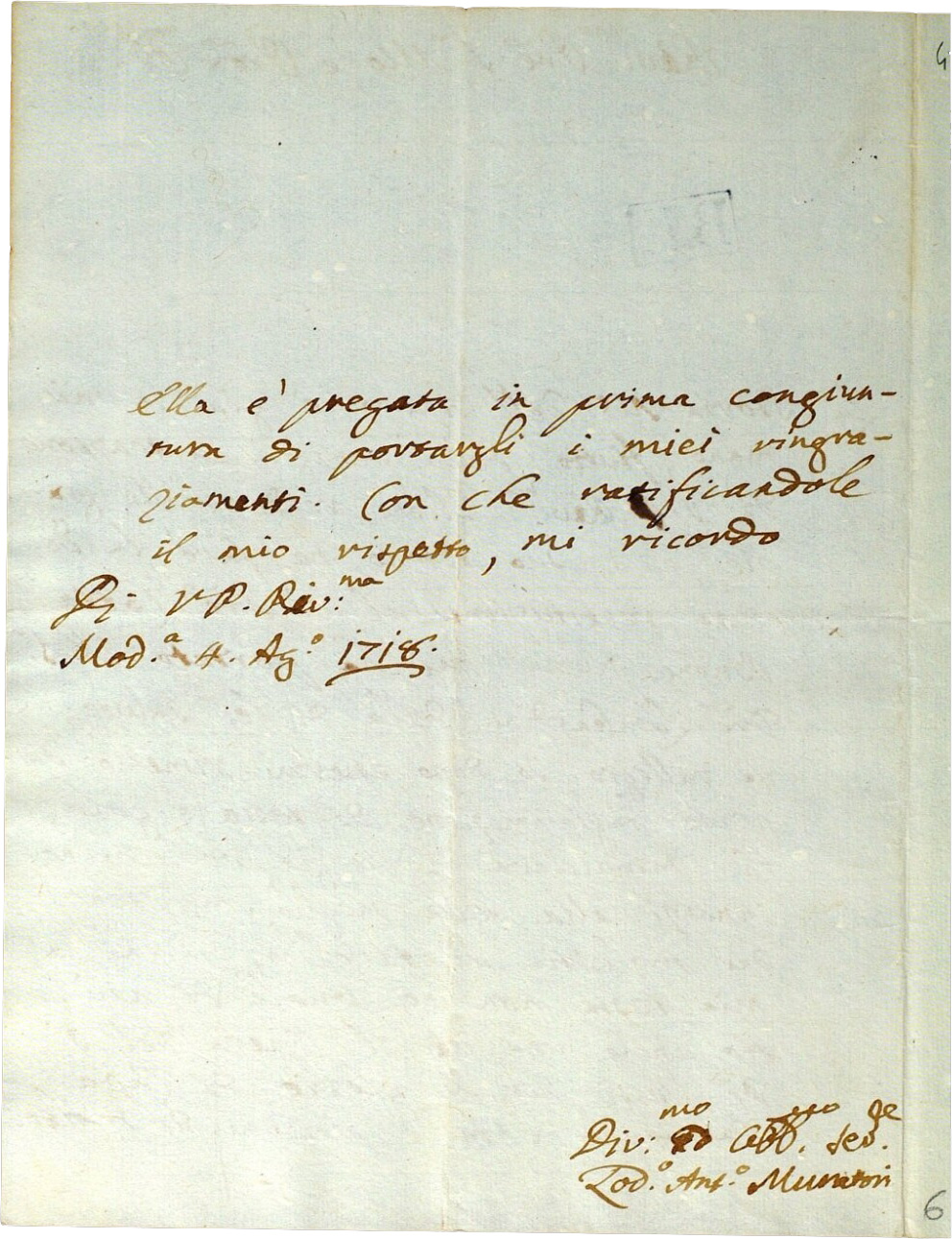} &
\includegraphics[width=0.15\linewidth]{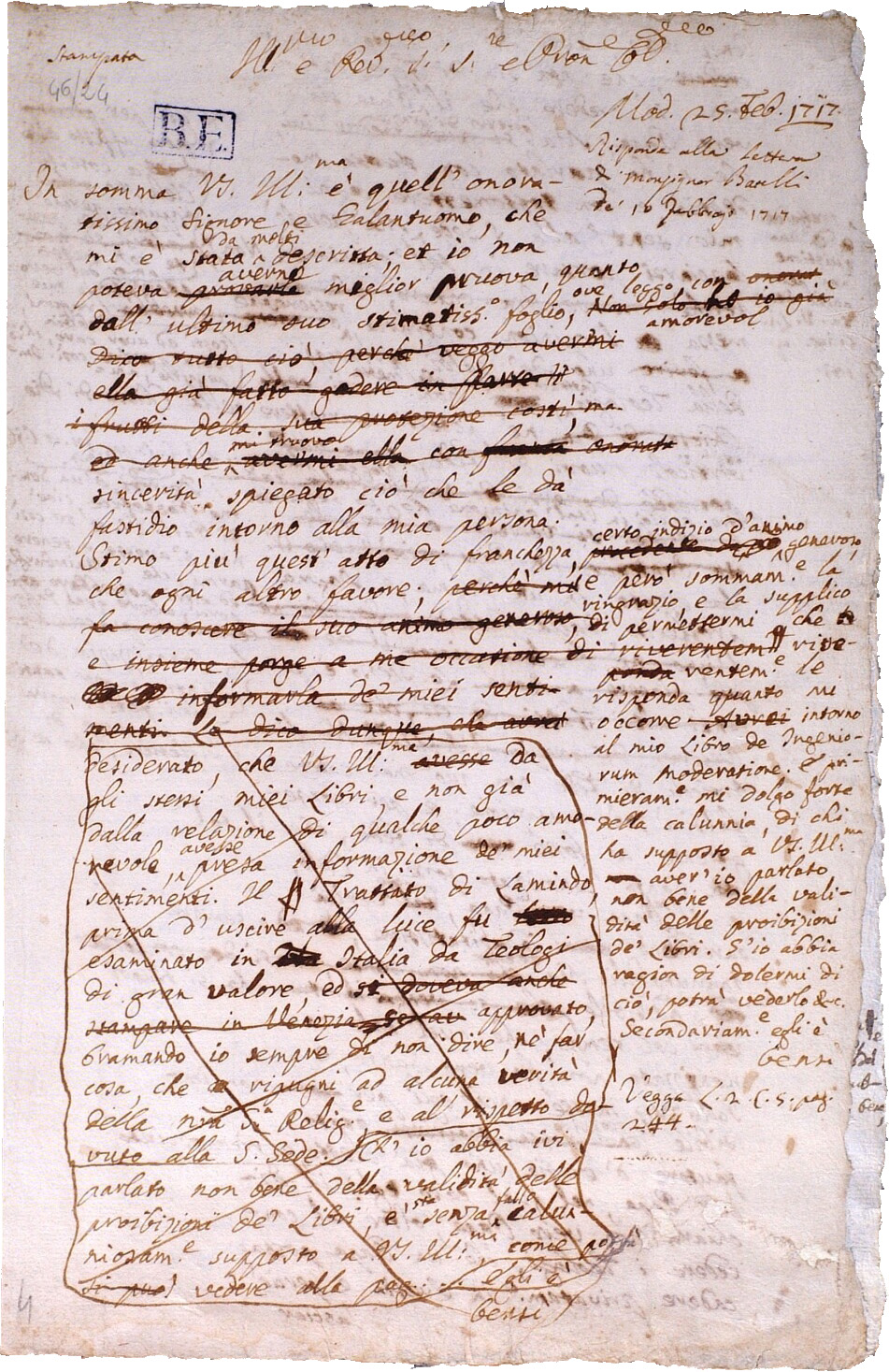} &
\includegraphics[width=0.15\linewidth]{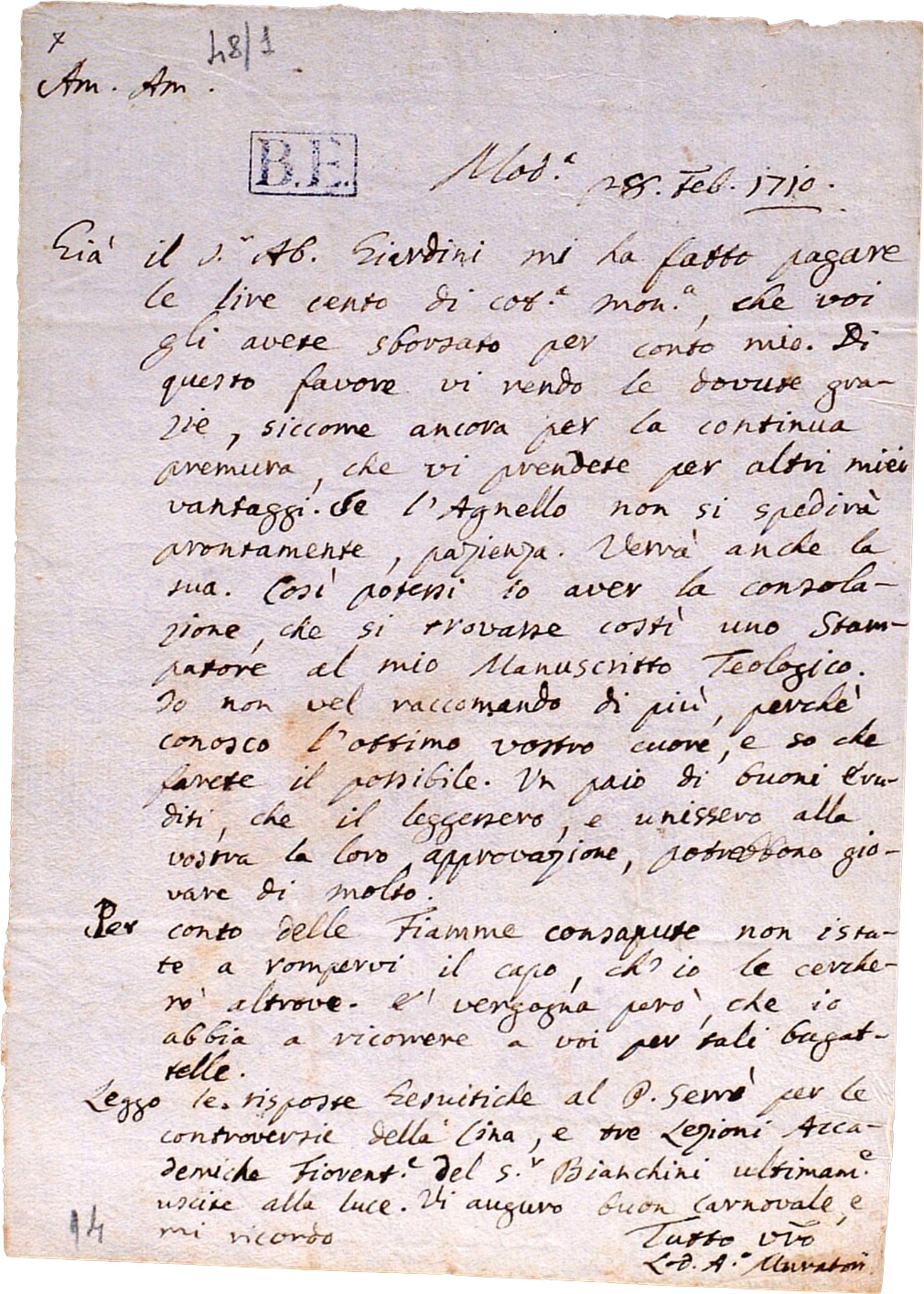} \\
\rotatebox{90}{\parbox[t]{1.4in}{\hspace*{\fill}\textbf{1720-1730}\hspace*{\fill}}}  & 
\includegraphics[width=0.15\linewidth]{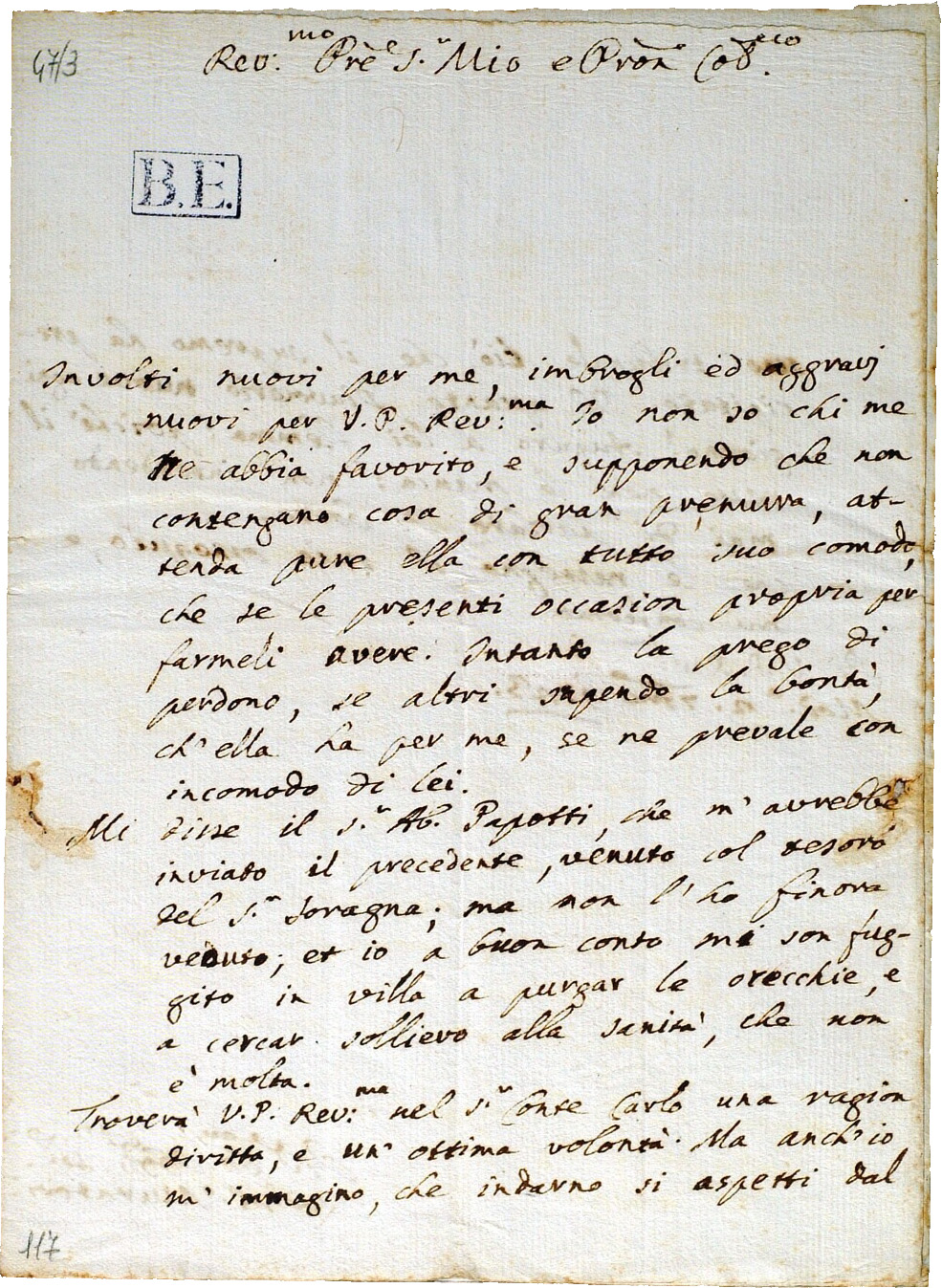} & 
\includegraphics[width=0.15\linewidth]{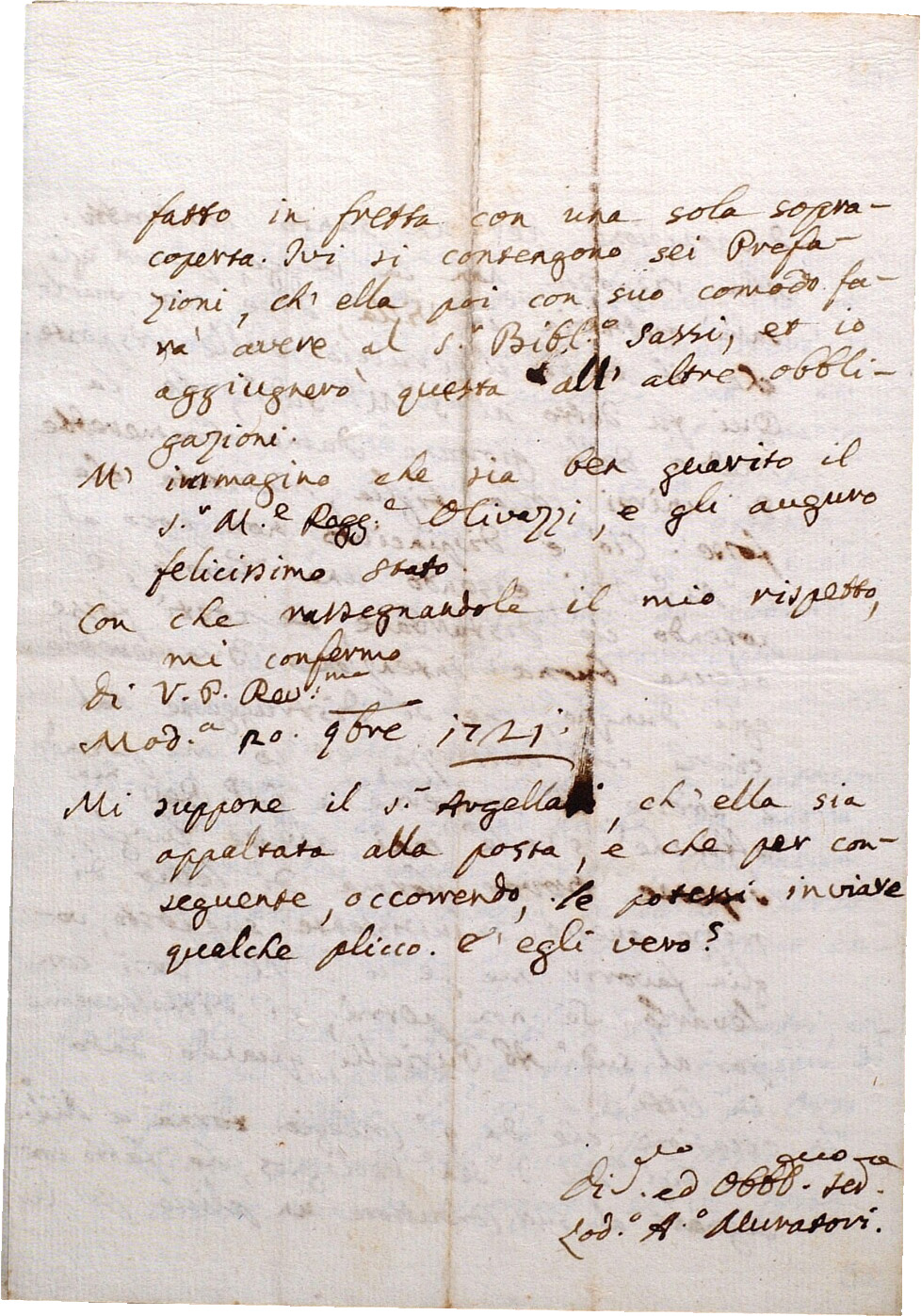} & 
\includegraphics[width=0.15\linewidth]{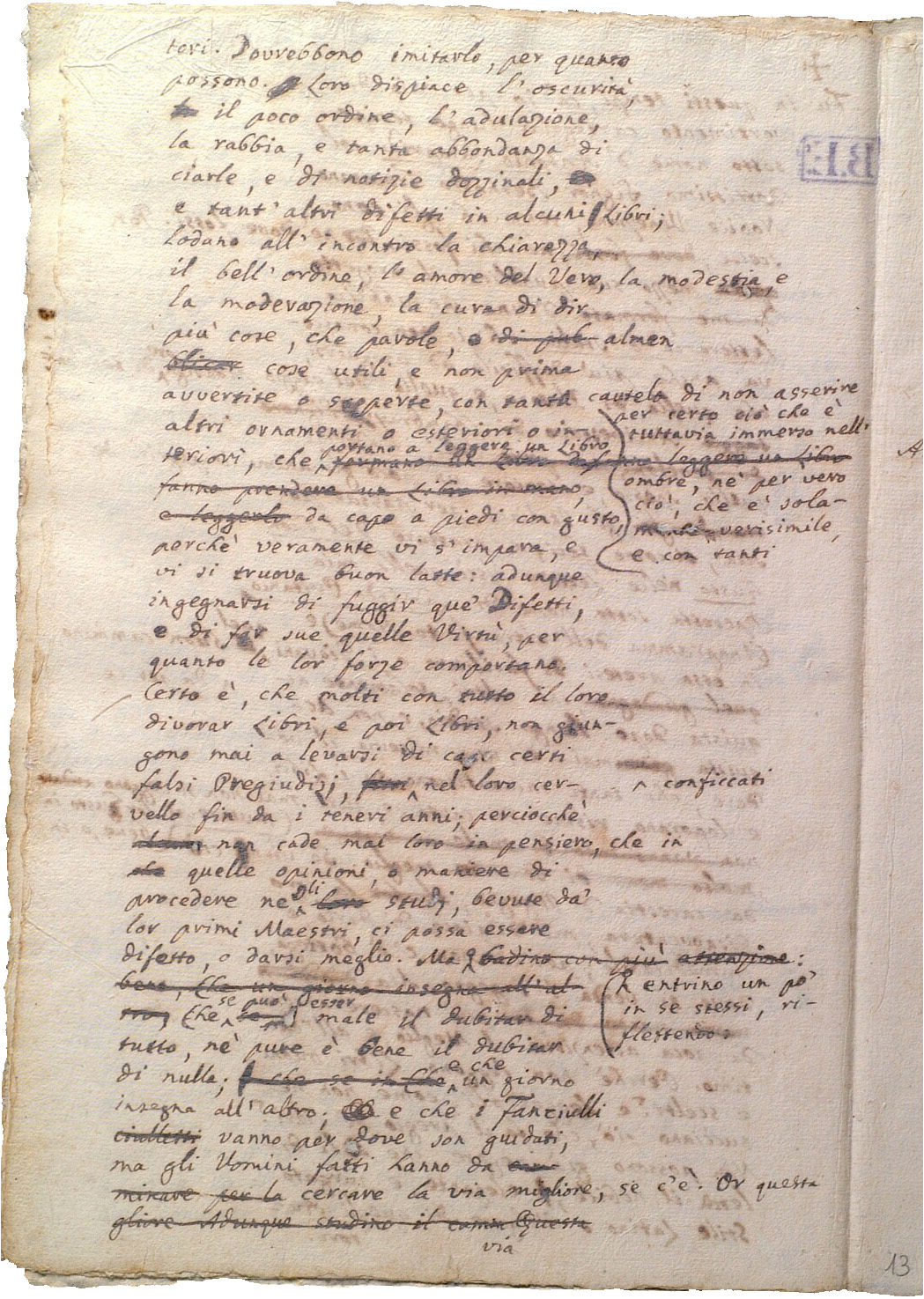} &
\includegraphics[width=0.15\linewidth]{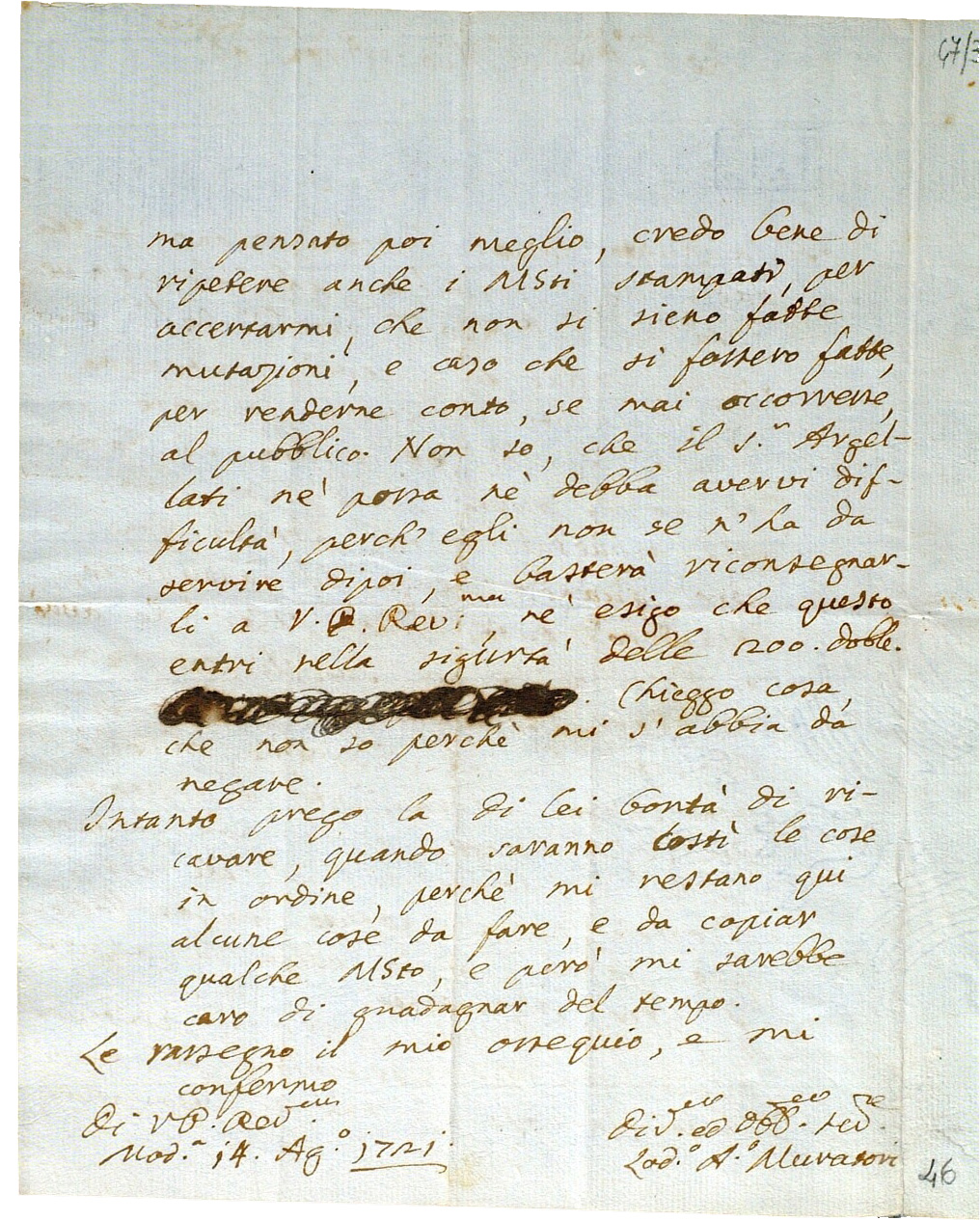} &
\includegraphics[width=0.15\linewidth]{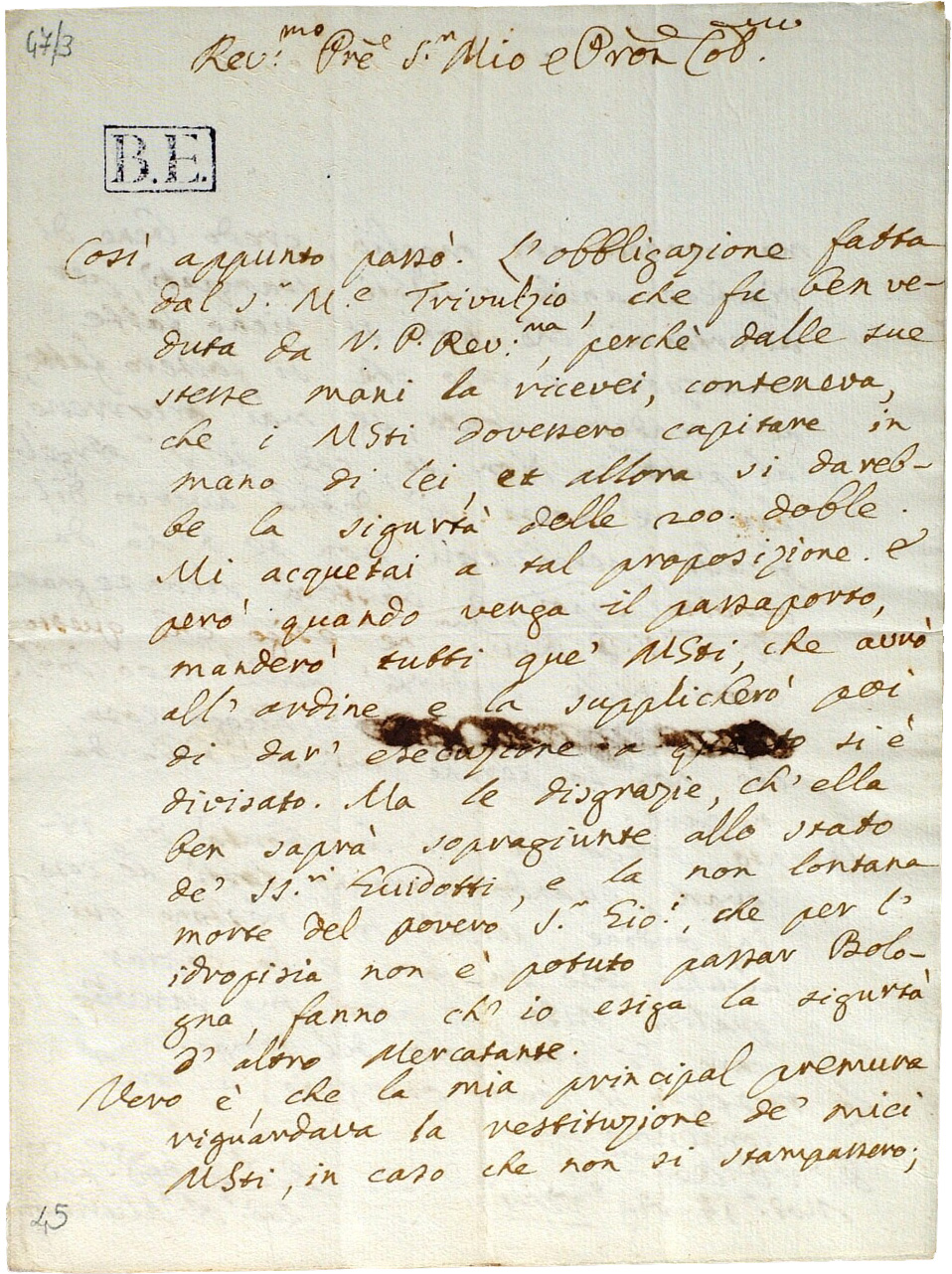} \\
\rotatebox{90}{\parbox[t]{1.35in}{\hspace*{\fill}\textbf{1730-1740}\hspace*{\fill}}}  & 
\includegraphics[width=0.15\linewidth]{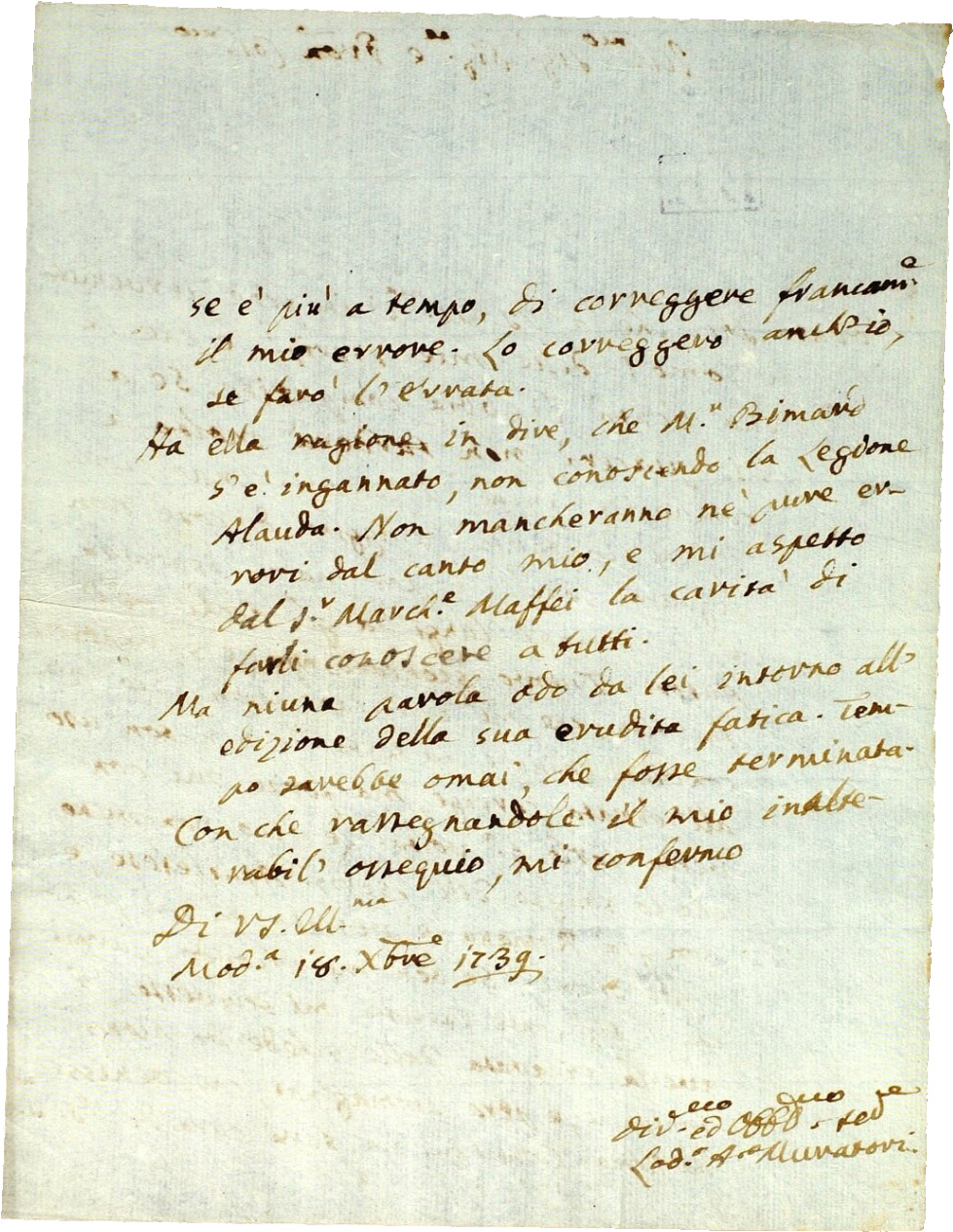} & 
\includegraphics[width=0.15\linewidth]{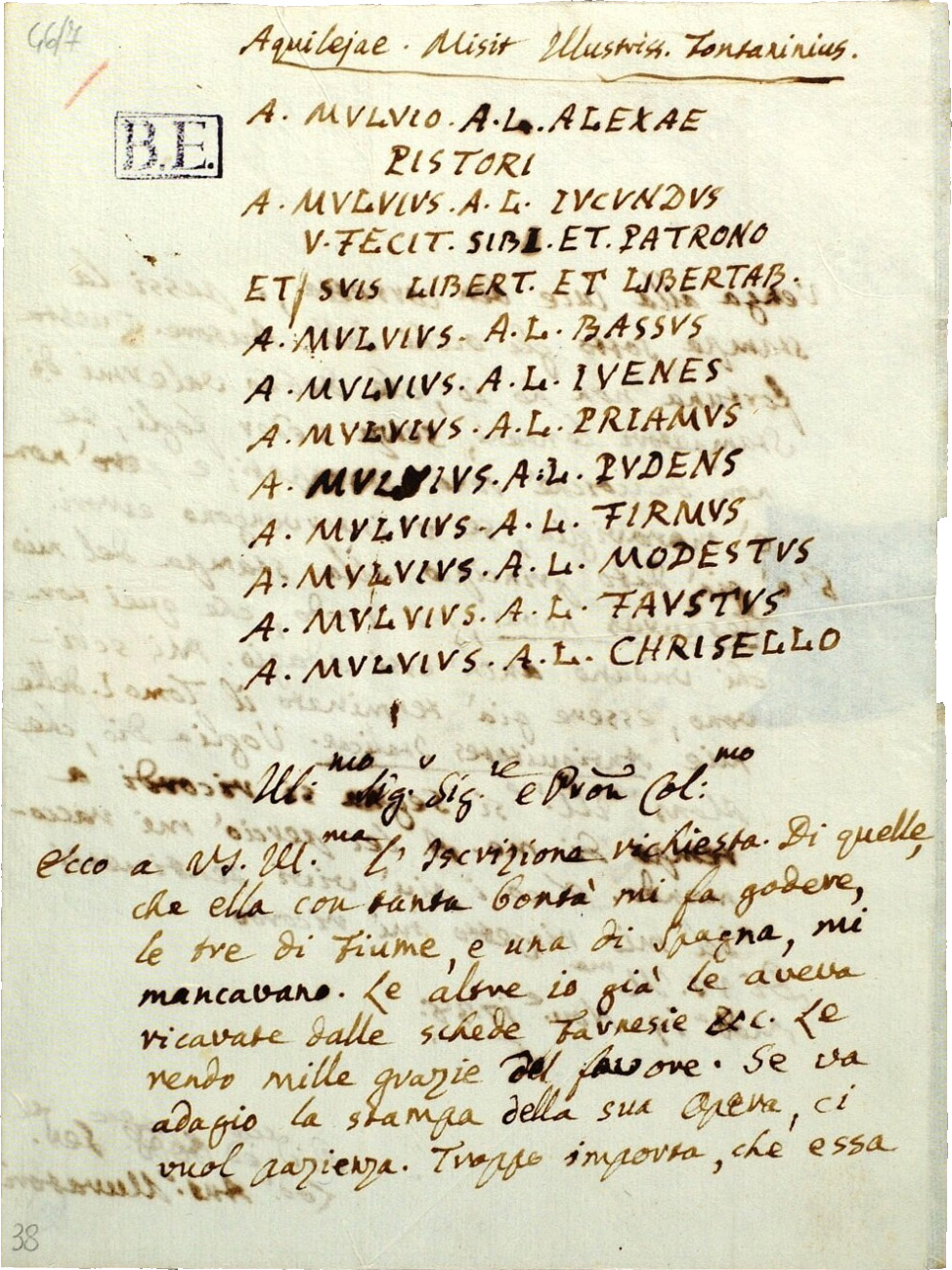} & 
\includegraphics[width=0.15\linewidth]{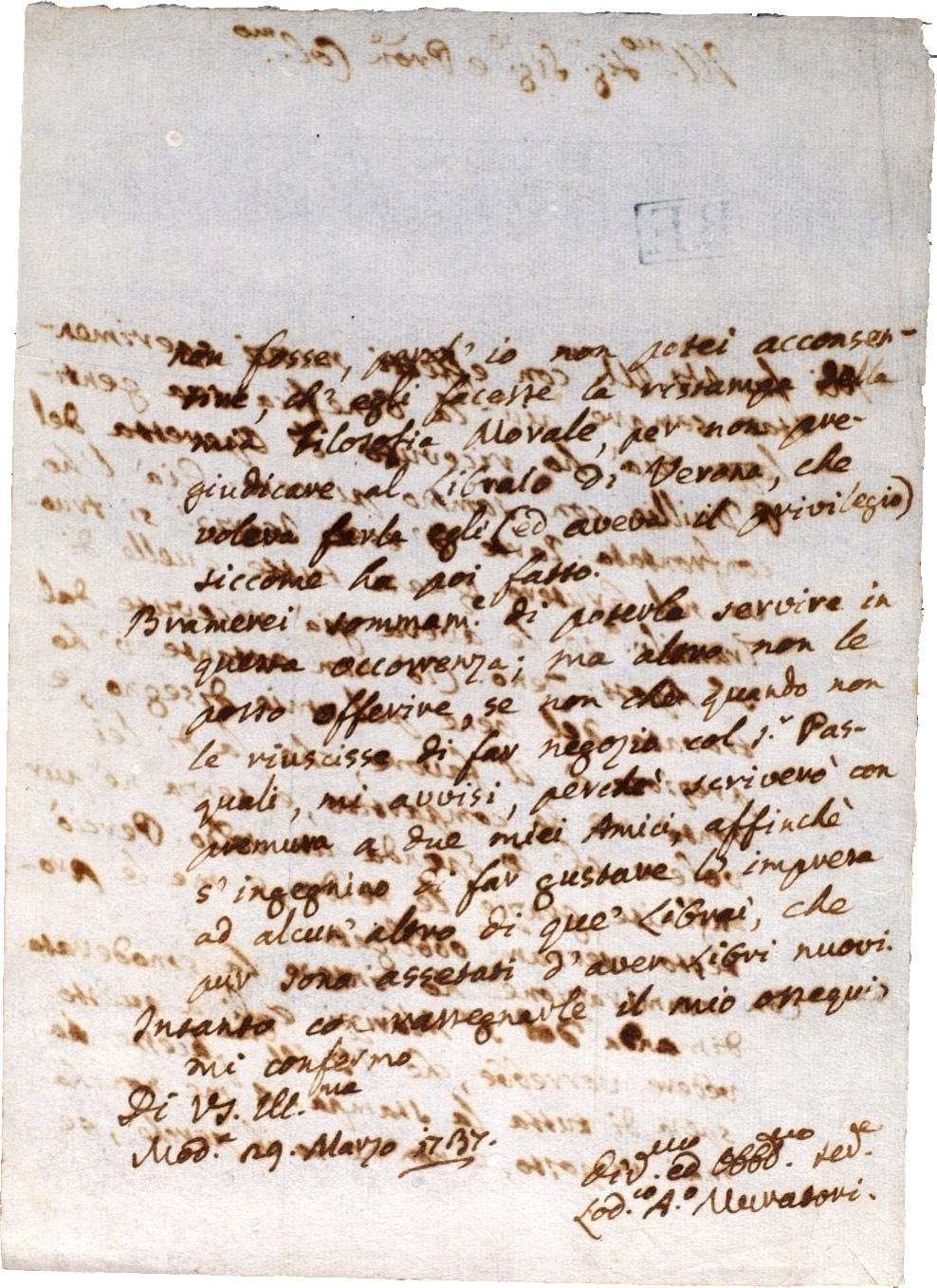} & 
\includegraphics[width=0.15\linewidth]{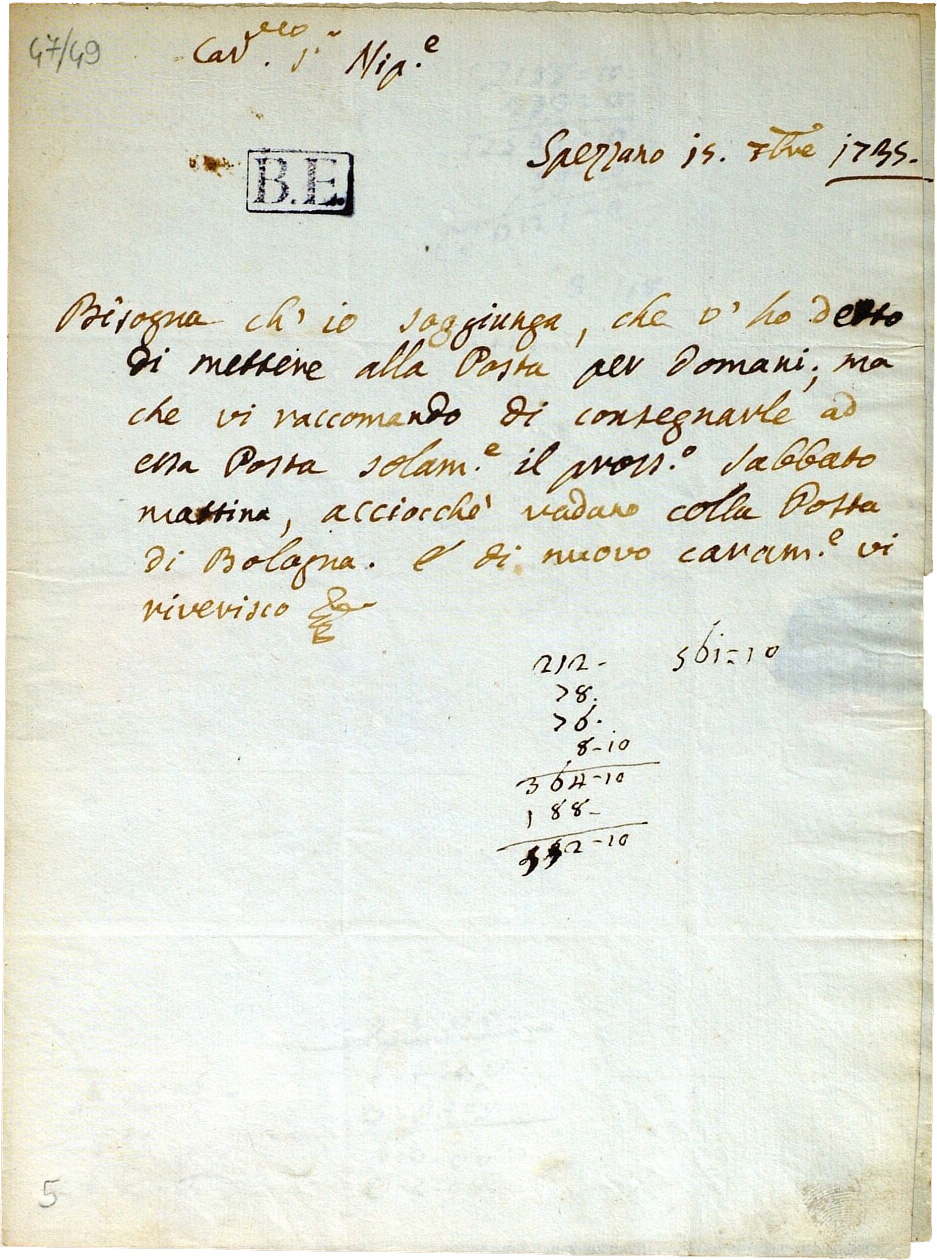} &
\includegraphics[width=0.15\linewidth]{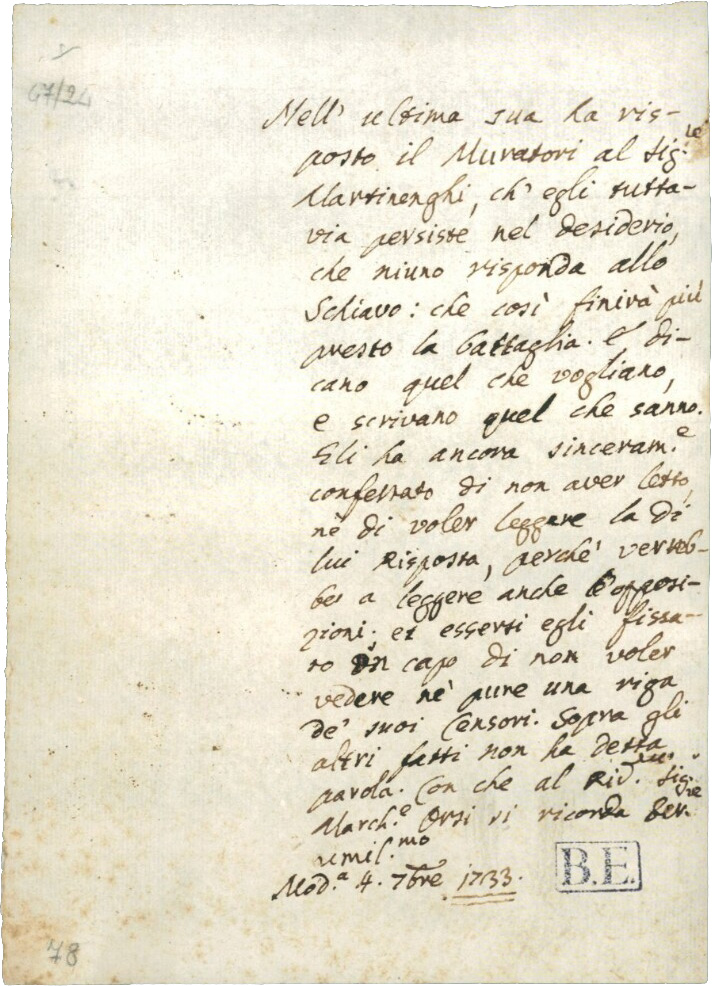} \\
\rotatebox{90}{\parbox[t]{1.4in}{\hspace*{\fill}\textbf{1740-1750}\hspace*{\fill}}}  & 
\includegraphics[width=0.15\linewidth]{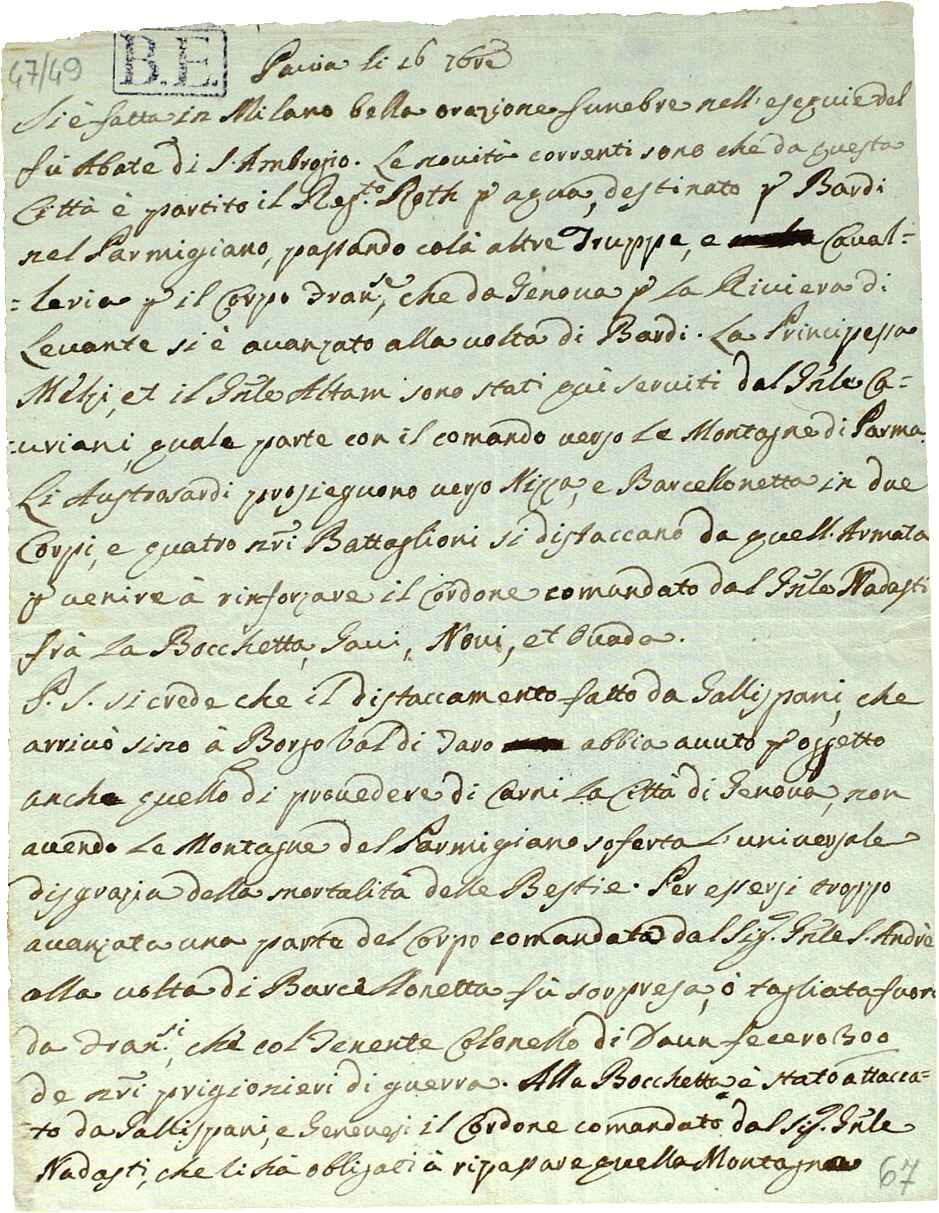} &
\includegraphics[width=0.15\linewidth]{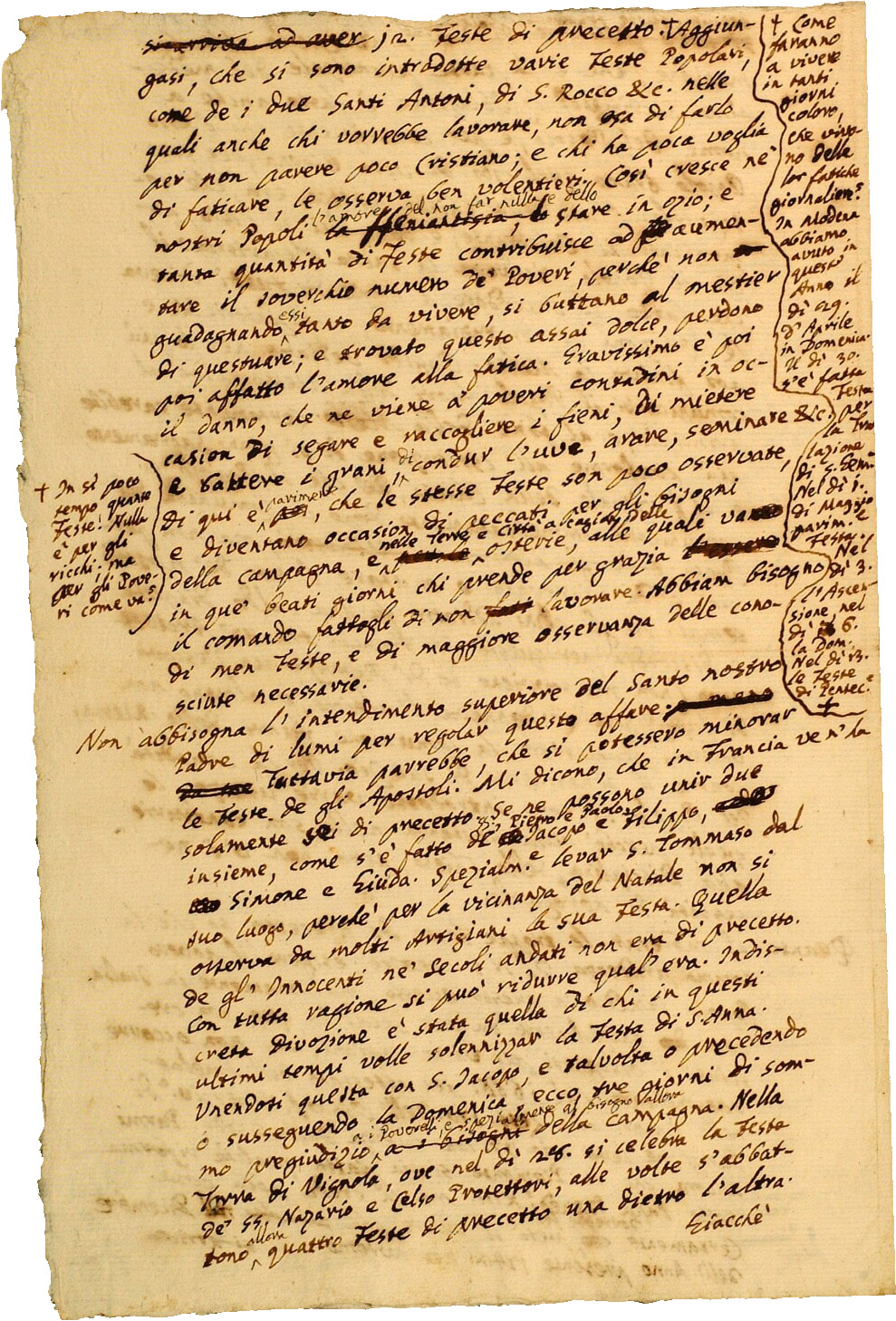} &
\includegraphics[width=0.15\linewidth]{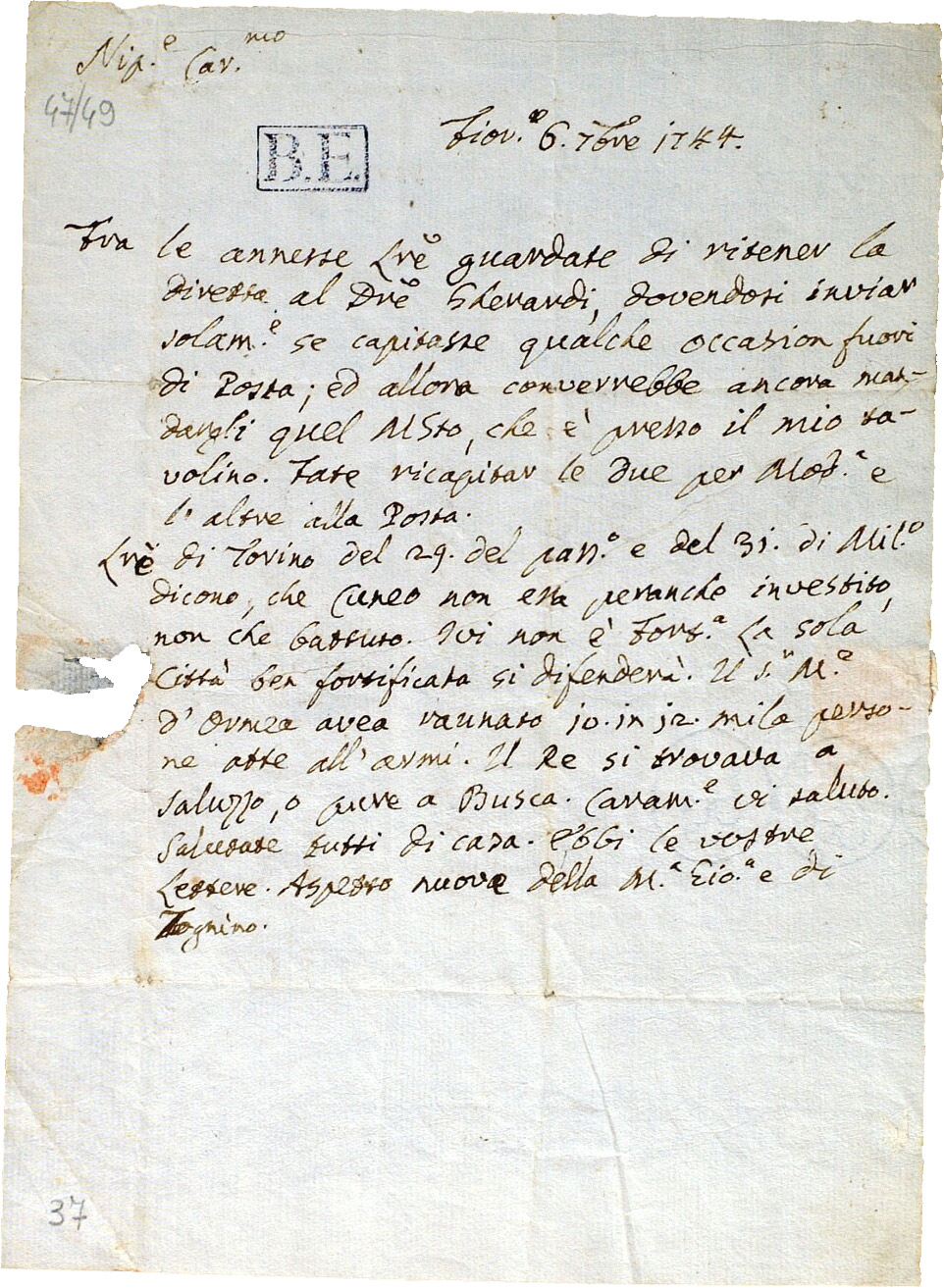} &
\includegraphics[width=0.15\linewidth]{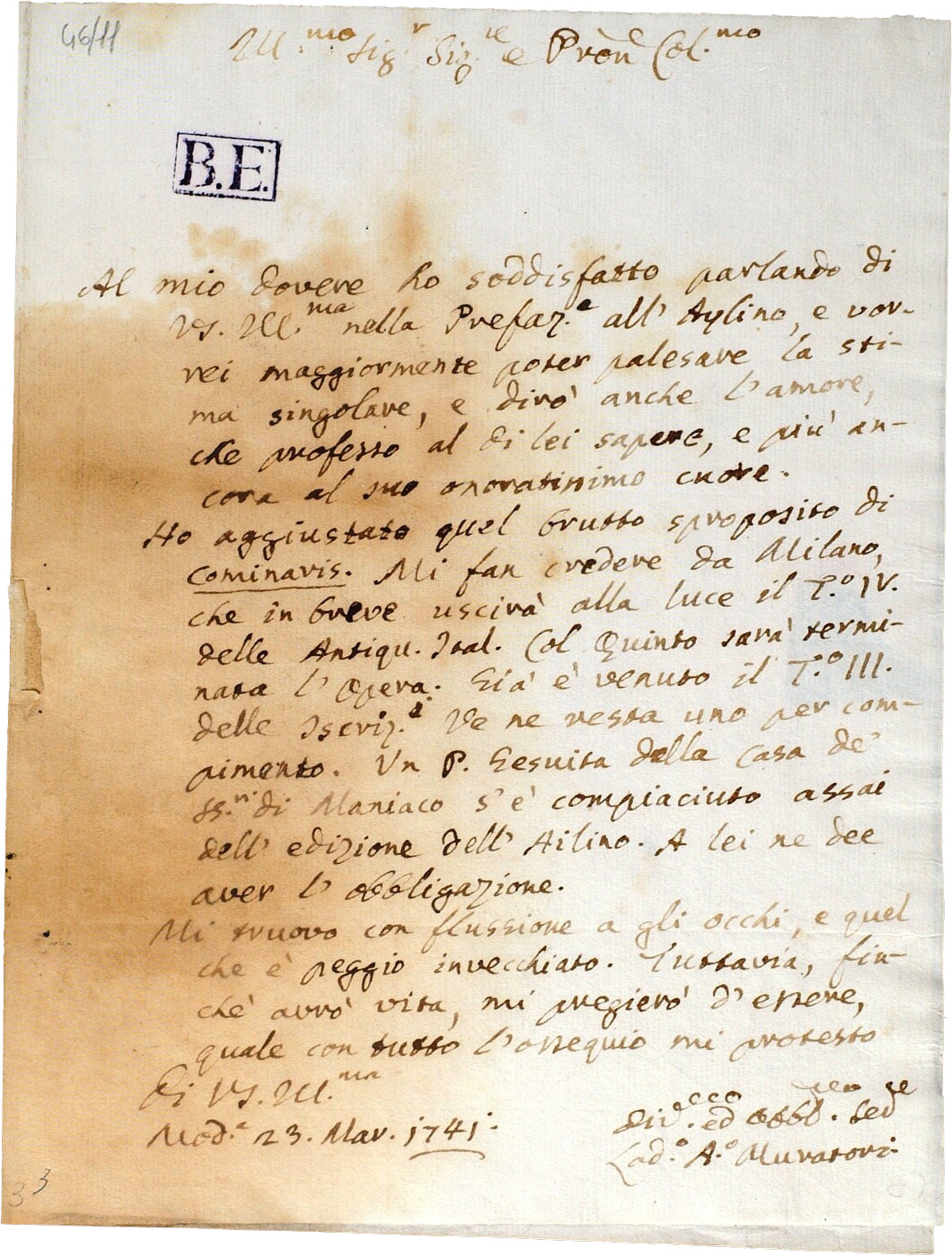} &
\includegraphics[width=0.15\linewidth]{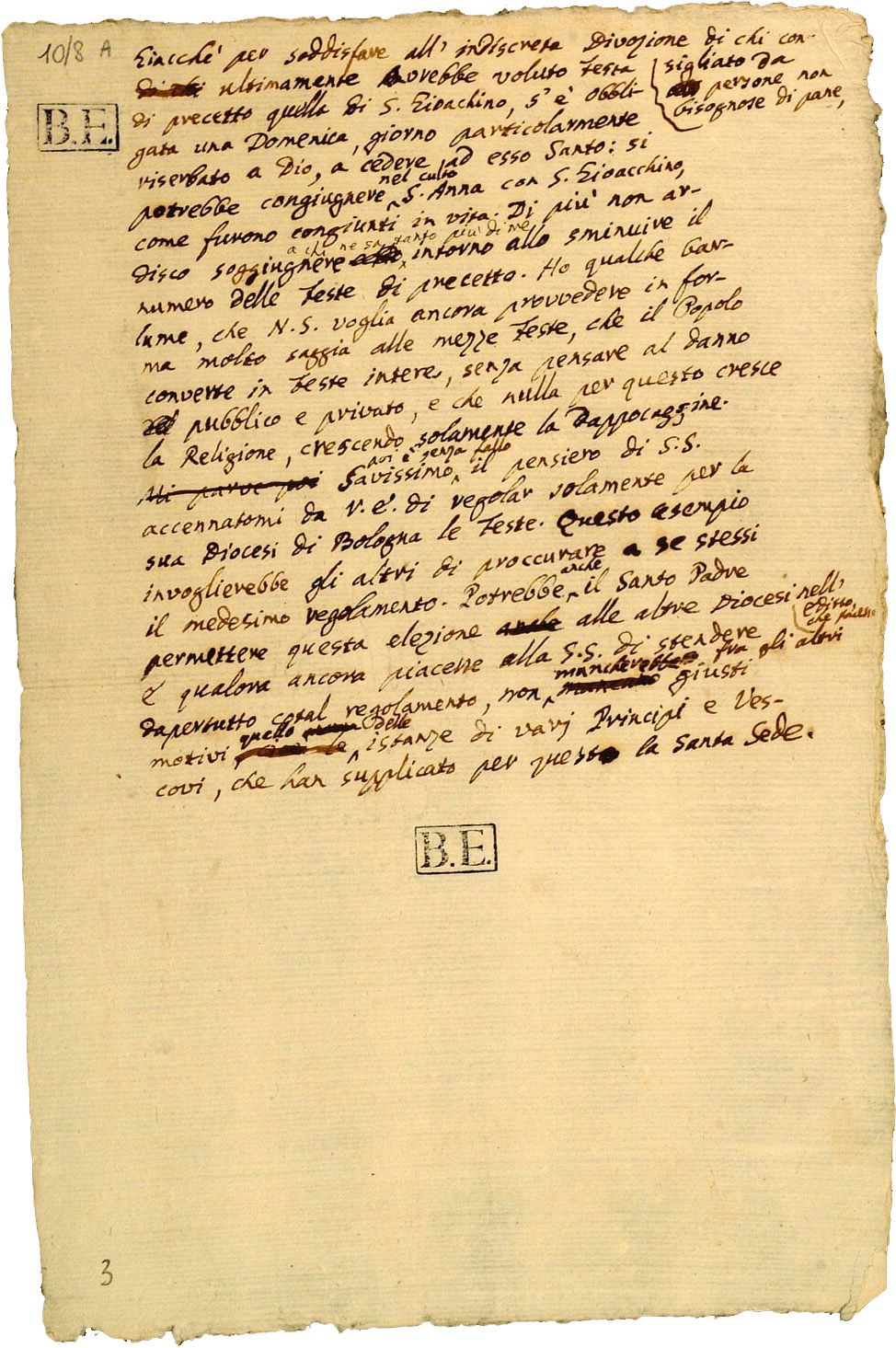} \\
\end{tabular}
}
\caption{Qualitative examples of pages written in different decades contained in the LAM dataset.}
\label{fig:variability}
\vspace{-0.2cm}
\end{figure*}

\bibliographystyle{IEEEtran}
\bibliography{bibliography}